\documentclass[letterpaper,twocolumn,10pt]{article}
\usepackage{usenix2019_v3}
\usepackage{float}
\usepackage{booktabs}
\usepackage{tikz}
\usepackage{subfig}
\usepackage{enumitem}
\usepackage{graphicx}
\usepackage{amssymb,amsmath,amsthm, graphicx,mathalfa,epsfig}
\usepackage{wrapfig}
\usepackage[linesnumbered,ruled]{algorithm2e}
\usepackage{algorithmicx}
\usepackage[noend]{algpseudocode}
\usepackage{svg}
\usepackage{array}
\usepackage{multicol,lipsum}
\usepackage{diagbox}
\theoremstyle{definition}
\usepackage{filecontents}

\theoremstyle{remark}

\begin{document}

\date{}

\title{\Large \bf On the Robustness of the Backdoor-based Watermarking \\in Deep Neural Networks}

\author{
{\rm Masoumeh \ Shafieinejad}\\
University of Waterloo
\and
{\rm Jiaqi Wang}\\
University of Waterloo
\and
{\rm Nils Lukas}\\
University of Waterloo
\and
{\rm Xinda Li}\\
University of Waterloo
\and
{\rm Florian Kerschbaum}\\
University of Waterloo}

\maketitle

\begin{abstract}
Obtaining the state of the art performance of deep learning models imposes a high cost to model generators, due to the tedious data preparation and the substantial processing requirements. To protect the model from unauthorized re-distribution, watermarking approaches have been introduced in the past couple of years. We investigate the robustness and reliability of state-of-the-art deep neural network watermarking  schemes. We focus on \emph{backdoor-based watermarking} and propose two -- a black-box and a white-box -- attacks that remove the watermark. Our black-box attack steals the model and removes the watermark with minimum requirements; it just relies on public unlabeled data and a black-box access to the classification label. It does not need classification confidences or access to the model's sensitive information such as the training data set, the trigger set or the model parameters. The white-box attack, proposes an efficient watermark removal when the parameters of the marked model are available; our white-box attack does not require access to the labeled data or the trigger set and improves the runtime of the black-box attack up to seventeen times. We as well prove the security inadequacy of the backdoor-based watermarking in keeping the watermark undetectable by proposing an attack that detects whether a model contains a watermark. Our attacks show that a recipient of a marked model can remove a backdoor-based watermark with significantly less effort than training a new model and some other techniques are needed to protect against re-distribution by a motivated attacker.
\end{abstract}

\section{Introduction}
Deep neural networks  (DNNs) have been successfully deployed in various applications; ranging from speech \cite{speech1,speech2,speech3} and image \cite{img1,img2,img3} recognition to natural language processing \cite{NLP1,NLP2,NLP3,NLP4} and more. The task of generating a model is computationally expensive and also requires a considerable amount of training data that has undergone a thorough process of preparation and labelling. The task of data cleaning is known to be the most time consuming task in data science~\cite{Dataclean}. According to the 2016 data science report, conducted by CrowdFlower, provider of a ``data enrichment" platform for data scientists, reveals that data scientists spend around $80\%$ of their time on just preparing and managing data for analysis~\cite{CFreport}. This enormous investment on preparing the data and training a model on it is however at an immediate risk, since the model can be easily copied and redistributed once released. To protect the model from unauthorized re-distribution, watermarking approaches have been introduced, inspired by wide deployment of watermarks in multimedia \cite{DWM1,DMW2,DWM3}. Watermarking approaches for DNNs lie in two broad categories: white-box and black-box watermarking. Black-box watermarking can be verified more easily than white-box watermarking; as in the former verification only requires API access to the service using the stolen model to verify the ownership of the deep learning model, while the latter requires model owners to access all the parameters of models in order to extract the watermark. Furthermore, black-box watermarking is advantageous over white-box watermarking as it is more likely to be resilient against statistical attacks~\cite{Florian}. 
\begin{figure}
    \centering
    \includegraphics[width=0.9\linewidth]{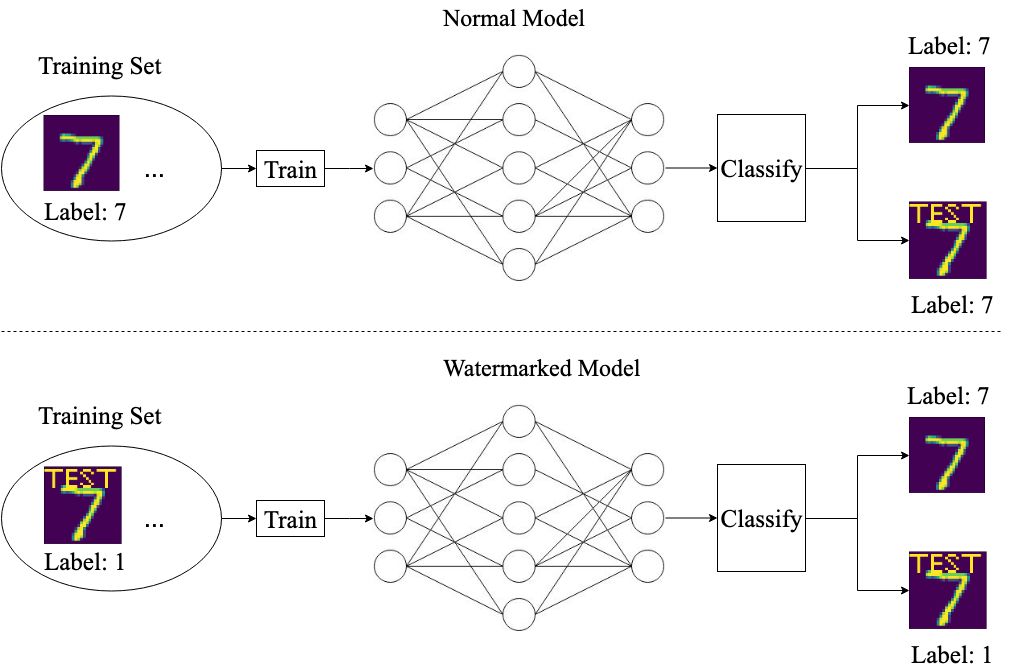}
    \caption{\label{fig:hlevel} A schematic illustration of the backdoor-based watermarking in neural networks}
\end{figure}
In this work
, we investigate recent black-box watermarking approaches proposed in \cite{usenix,asiaccs,guo}, these approaches each introduce a (some) variant (s) of backdoor-based watermarking to protect model ownership. Backdoors or neural trojans, \cite{backatt1,backatt2,backatt3}, originally are the terms for a class of attacks against the security of deep learning when an entity outsources the learning for model computation to another untrusted but resourceful party. The party can train a model that performs well on the requested task, while its embedded backdoors lead to targeted misclassifications when encountering a particular trigger in the input. The idea of ``turning weakness to a strength"~\cite{usenix}, launched a new line of work suggesting using backdoors for ownership protection \cite{usenix,asiaccs,guo}. The motivation behind the research on this topic is to use the trigger to embed a \emph{signature} of the model owner, as shown in Figure \ref{fig:hlevel}. The trigger in the input can take one of the following forms: embedded content representing a logo of the owner \cite{asiaccs,guo}, a pre-defined noise pattern in the inputs \cite{asiaccs} or a set of particular inputs acting as a secret \emph{key} set \cite{asiaccs,usenix}. We investigate whether using backdoors in DNNs brings sufficient robustness for watermarking the models. We introduce two attacks (black-box and white-box) on the aforementioned backdoor-based watermarking schemes, and show that these watermarks are removable in both attacks. Our attacks require neither the original labeled training data, nor the backdoor embedded in the model. We have two goals: i) removing the watermark, while ii) providing the same functionality as the original marked model. These goals are achievable by both our black-box or white-box attack; the black-box attack has the minimum requirements on information available to the attacker, while the white-box attack provides more efficiency. Our black-box attack sheds more light on model stealing and how it exploits the fact that backdoors do not necessarily transfer among models if the models are not trained on the exact same data. Stealing the functionality of models was well generalized by Orekondy et al. \cite{Knockoff}, with less requirements than similar works in the literature \cite{PRADA,Steal2}. 
We use a similar approach to Orekondy et al. in our black-box attack, with two differences: i) we query the target model with data of a similar distribution to the model's training data rather than random images, and ii) we just need the final label for the queries instead of their probability vectors. Limiting the availability of the probability vectors to the availability of the final labels is a proposed defence against the previous model stealing attacks \cite{Steal2}. Our results show that an attacker can steal a model that is trained by a resourceful party and remove the watermark without losing accuracy or any need to undergo an extensive effort of data preparation as required for the watermarked model. It is worth noting existing approaches for removing backdoors are not applicable to removing watermarks. Neural-Cleanse \cite{Neural-Cleanse} requires access to a set of correctly labeled samples to detect the backdoor and only works for one type of backdoor used in watermarking
-- a small patch on the images. Fine-Pruning \cite{Prune} removes backdoors by pruning redundant neurons that are less useful for the main classification; this technique drops the model accuracy rapidly for some models \cite{Knockoff}. We propose a white-box attack that covers a wider range of backdoors with up to seventeen times efficiency than our black-box attack. Our white-box attack does not affect model accuracy and our black-box attack achieves accuracy close to the marked model.

\subsection{Our Contributions} We propose three main contributions in this work:  (i) We introduce our black-box attack that removes the embedded watermark in backdoor-based watermarking schemes. Our attack solely relies on publicly available information, i.e. no labeled data, and successfully removes the watermark from the neural network without requiring any access to the network parameters, the classification probability vector, the backdoor embedded as the watermark, or the training data. The performance and accuracy of the stolen model in this attack is very close to the watermarked model. (ii) We present a white-box attack for scenarios that we are guaranteed access to the model parameters. Benefiting from the additional information, our white-box attack offers a(n) (up to seventeen times) faster version of the black-box attack, without affecting the watermarked model's accuracy. 
(iii) We as well present our property inference attack that fully distinguishes watermarked neural networks from an unmarked ones. While our first two contributions show that backdoor-based watermarking schemes do not satisfy the unremovability property, our last contributions shows these schemes are also inadequate to fulfill undetectability. Exploiting these two security vulnerabilities in combination, empowers the attacker to remove the the watermarks completely and efficiently. Therefore, different protections techniques against these attacks are necessary.

\subsection{Paper Organization} The rest of this paper is organized as follows: We provide formal definitions for deep neural networks and backdoor-based watermarking in Section \ref{Backdoor-based_WM} in addition to describing the security vulnerability in the schemes. Subsequently in Section \ref{Attacks}, we introduce our black-box and white-box attacks for watermark removal. We as well propose an attack that detects the presence of a watermark in a given model. Finally in Section \ref{Experiments}, we present the results of the experiments that confirm a successful watermark removal. In Section \ref{Related_Work}, we situate our work in the current body of research.

\section{Backdoor-based Watermarking}\label{Backdoor-based_WM}
Backdooring enables the operator to train a model that deliberately outputs specific (incorrect) labels; backdoor-based watermarking schemes use this property to design trigger sets to watermark the DNN. The intuition in black-box watermarking is to exploit the generalization and memorization capabilities of deep neural networks to learn the patterns of an embedded trigger set and its pre-defined label(s). The pairs of learned patterns and their corresponding predictions will act as the keys for the ownership verification. As described in Section \ref{backdoor-based_WM_types}, we focus on three backdoor-based watermarking schemes proposed in the literature. Zhang et al. \cite{asiaccs} generate three different types of trigger sets for deep neural network models by: (a) embedding a content, e.g. adding a logo, into a sample of the original training data,  (b) adding static Gaussian noise as watermarks into a sample of the original training data, or  (c) embedding out-of-distribution data samples. Guo and Potkonjak \cite{guo} propose a content embedding approach (similar to the logo in \cite{asiaccs}) for watermarking in DNNs. Similar to the third watermark generation algorithm in \cite{asiaccs}, Adi et al.~\cite{usenix} suggest embedding a backdoor set of abstract images in it. In what follows, we provide a formal definition for learning process in neural networks, we also formally describe backdoor-based watermarking in DNN and elaborate on the schemes introduced in \cite{asiaccs,usenix,guo}. We end the section by pointing out the security vulnerabilities of the schemes.

\subsection{Definitions and Models}\label{Definitions}
We follow the notations of \cite{usenix} throughout this paper to introduce our attack accordingly. In order to train a neural network, we initially require some objective ground-truth function $f$. The neural network consists of two algorithms: training and classification. In training, the network tries to learn the closest approximation of $f$. Later, during the classification phase the network utilizes this approximation to perform prediction on unseen data. 
\begin{figure}[h!]
    \centering
    \includegraphics[width=0.9\linewidth]{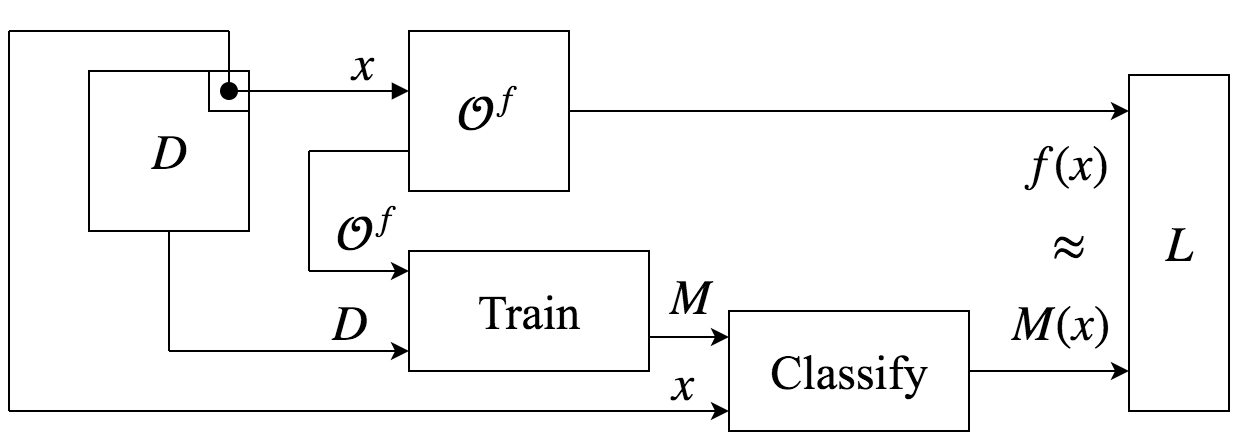}
    \caption{A high-level illustration of the learning process}
    \label{fig:ML}
\end{figure}
Formally, the input to the neural network is represented by a set of binary strings: $D \subseteq \{0,1\}^*$, where $|D|=\Theta (2^n)$, with $n$ indicating the input length. The corresponding labels are represented by $L \in \{0,1\}^* \bigcup \{\bot \}$ and $|L|= \Omega (p (n))$ for a positive polynomial $p (.)$; with the symbol $\bot$ showing the undefined classification for a specific input. The ground-truth function $f: D \rightarrow L$, assigns labels to inputs. Also, for $\Bar{D}$ the set of inputs with defined ground-truth labels, $\Bar{D}= \{x \in D|f (x) \neq \bot\}$, the algorithms' access to $f$ is granted through an oracle $\mathcal{O}^f$. Hence, the learning process illustrated in Figure \ref{fig:ML} of~\cite{usenix}, consists of the following two algorithms:
\begin{itemize}
    \item Train ($\mathcal{O}^f$): a probabilistic polynomial-time algorithm that outputs a model $M \subseteq \{0,1\}^{p (n)}$ 
    \item Classify ($M,x$): a deterministic polynomial-time algorithm that outputs a label $M (x) \in L \backslash \{\bot\}$ for each input $x \in D$
\end{itemize}
The metric $\epsilon-$accuracy evaluates the accuracy of the algorithm pair  (Train, Classify). In an $\epsilon-$accurate algorithm the following inequality holds: $Pr[Classify  (M,x) \neq f(x) | x \in \Bar{D}] \leq \epsilon$; the probability is taken over the randomness of Train, with the assumption that the  ground-truth label is available for those inputs. 
\subsection{Backdoor-based Watermarking in DNNs}
\begin{figure}[h!]
    \centering
    \includegraphics[width=1\linewidth]{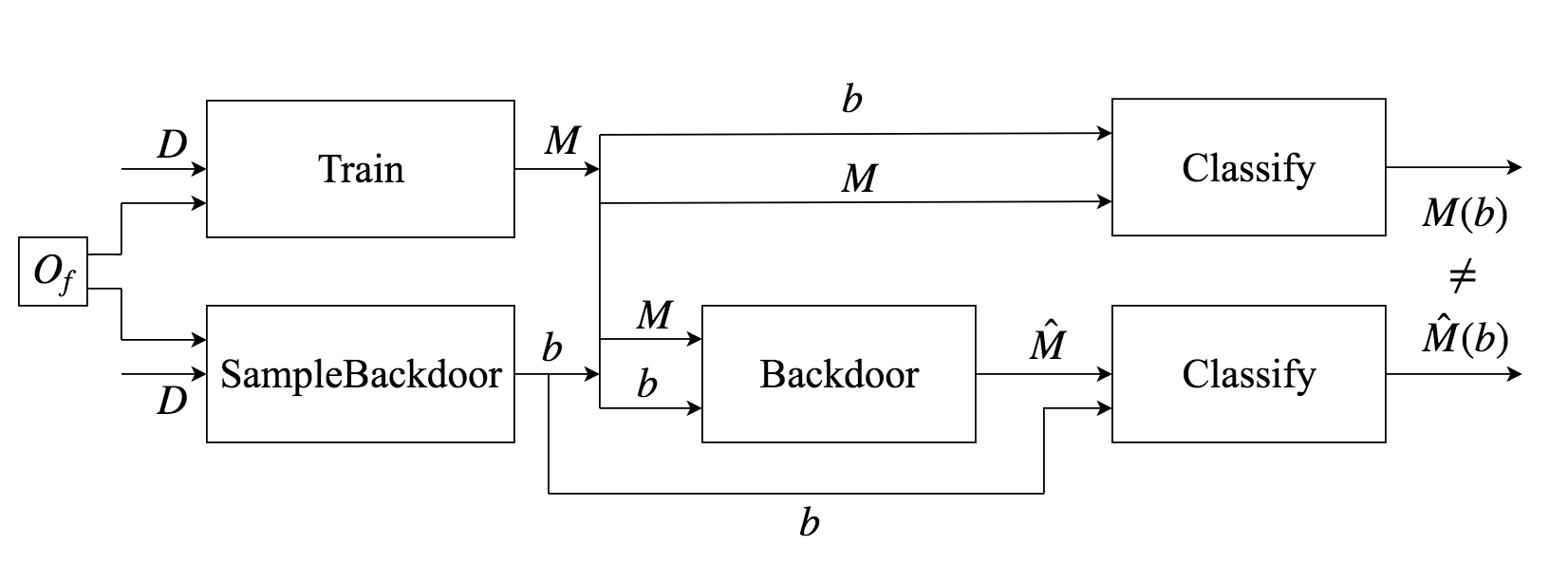}
    \caption{A schematic illustration of the backdooring process}
    \label{fig:backdooring}
\end{figure}

Backdooring teaches the machine learning model to output \emph{incorrect} but valid labels $T_L :T \rightarrow L \backslash \{\bot\}; x \mapsto T_L (x) \neq f (x)$ to a particular subset of inputs $T \subseteq D$, namely the trigger set. The pair $b= (T, T_L)$ forms the backdoor for a model. A randomized algorithm called $SampleBackdoor$ generates the backdoors $b$. $SampleBackdoor$ needs access to the oracle $\mathcal{O}^f$ and works closely with the original model. The complete backdooring process is illustrated in Figure \ref{fig:backdooring}~\cite{usenix}.
\\Formally presenting, backdoor ($\mathcal{O}^f, b , M$) is a PPT algorithm that on inputs oracle to $f$, the backdoor $b$ and a model $M$, outputs a model $\hat{M}$. The backdoor model $\hat{M}$ is required to output particular \emph{incorrect} (regarding $f$) labels for the inputs from the trigger set and correct ones for other inputs. In other words, the following two inequalities must always hold for a backdoored model $\hat{M}$:
\begin{itemize}
    \item $Pr_{x \in \Bar{D}\backslash T}$\footnote{From this point on, we assume $\Bar{D}=D$, without loss of generality} $ [Classify(\hat{M},x) \neq  f (x)] \leq \epsilon$
    \item $Pr_{x \in T}[Classify  (\hat{M},x) \neq T_L (x)] \leq \epsilon$
\end{itemize}
To watermark an ML model using the backdooring process, the algorithm $MModel ()$ is used. MModel consists of the following sub-algorithms:
\begin{enumerate}
    \item Generate $M \leftarrow$ $Train (\mathcal{O}^f$): Generates the original model on the training set, not that the trigger set is not included in training in this step. 
    \item Sample($mk, vk$) $\leftarrow$ KeyGen(): The watermarking schemes commits to the embedded backdoors $b$ and generates marking and verification keys from $b$. Since the details of generating $mk ,vk$ has no affect on our attack, we leave them as out of scope of this work. 
    \item Compute $\hat{M} \leftarrow Mark (M, mk)$: Computes the watermarked model, by embedding the backdoors $b$.
    \item Output $ (M, \hat{M}, mk, vk)$.
\end{enumerate}
The verification of a watermark is performed by the  algorithm $Verify (mk, vk, M)$. $Verify$ takes as input the marking and verification keys and a model $M$, and outputs a bit $b \in \{0,1\}$, indicating whether the watermark is present in the model $M$ or not. Formally,\\ ${Verify (mk, vk, M)}$ \\
\hspace*{4ex} =\ $\begin{cases}
    \text{1, } \text{ if } \sum_{x \in T}\mathbb{I}[Classify (M,x) \neq T_L{(x)}]- \frac{1}{|L|}|T| \leq \epsilon|T|\\
    \text{0, } \text{ otherwise.}\\
\end{cases}$
where $\mathbb{I}$[expr] is the indicator function that evaluates to 1 if expr is true and 0 otherwise. Note that, as we skip the commitment details, the marking key $mk$ in $Verify$ translates to the inputs $x$ in the trigger set $T$, and the verification key $vk$ refers to the corresponding labels $T_{L}$. Furthermore, the $\frac{1}{{|L|}}|T|$ comes from the assumption that the ground-truth label is undefined for the inputs of the trigger set $T$, for which we assume the label is random. Hence, we assume that for any $x \in T$, we have $Pr_{i \in L}[Classify (M,x)=i] =\frac{1}{|L|}$. As a result, it is expected that  $\frac{1}{{|L|}}|T|$ of the inputs fall into the backdoor label ``randomly". Hence, to verify the presence of the watermark in the model without a bias, we need to deduct this number from the classification result.    
\subsection{Backdoor-based Watermarking Schemes}\label{backdoor-based_WM_types}
We investigate the recent backdoor-based schemes in \cite{asiaccs,usenix,guo}. The watermark embedded in these schemes is one of the following three forms: Embedded Content, Pre-Specified Noise, and Abstract Images.
\begin{enumerate}[label=\alph*)]
    \item Content Embedded (Logo):
    This method \cite{guo,asiaccs} adds a fixed visual marker (e.g. a text) to a set of inputs, namely the watermark set. The inputs with this watermark all classify to a fixed label. 
    \item Pre-Specified Noise: 
    This method \cite{asiaccs}, adds a fixed instance of Gaussian noise as watermark to inputs. Similar to Content Embedded watermarking schemes, this approach maps the marked inputs to a fixed label.
    \item Abstract Images:
    In this category of watermarking, a set of abstract images~\cite{usenix}, or images that are not from the same distribution as the training data~\cite{asiaccs} is selected and labeled with pre-defined classes. 
\end{enumerate}
The are two main differences between the last watermarking and the first two, in Abstract Images: i) different subsets of the trigger set are mapped to different classes, not just a fixed one, and ii) the watermark is no more a pattern added to any input, it is a fixed set of inputs and labels. Hence, the watermark test set is exactly the same as the watermark train set. 
\subsection{Backdoor-based Watermarking - Security}\label{Security_Backdoor-based_WM}
We review the security claims in black-box watermarking as stated in \cite{asiaccs,guo,usenix}, and discuss how our attacks invalidate the presented claims.  (i) As stated in ~\cite{asiaccs}, Section 4: ``After embedding (the watermarks), the newly generated models are capable of ownership verification. Once they [the marked models] are stolen and deployed to offer AI service, owners can easily verify them [the watermarks] by sending watermarks as inputs and checking the service's output."  (ii) As well, authors in \cite{usenix} claim that their proposed scheme is persistent in the sense that ``It is hard to remove a backdoor, unless one has knowledge of the trigger set." However, our attacks show that the watermark is successfully removable -- hence, it is no longer verifiable -- and to perform so, the attacker does not require any knowledge of the trigger set. 
(iii) On a similar note to \cite{usenix}, Guo et al.~\cite{guo} claims that ``transferring learning is on the same order of magnitude as the cost of training, if not higher. With that much resources and expertise at hand, an attacker would have built a model on their own." This claim neglects the fact that the cost of data preparation for the original model is comparable with (or even higher than) the cost of model generation itself, consequently the attacker saves a considerable amount by stealing the model through queries.  (iv) Zhang et al.~\cite{asiaccs} refer to the results of \cite{Steal} to state that stealing a model using prediction APIs needs queries of significant size; e.g. $100k$, where $k$ is the number of model parameters in the particular example of two-layered neural network in \cite{Steal}. They conclude that as more complicated models with more parameters the attack would even need considerably more queries. Our experiments demonstrate that a successful model stealing attack on ResNet-32 network with 0.46 million of parameters \cite{ResNet}, only needs to query the API, 50,000 times and does not require access to the probability vector. (v) Furthermore, Adi et al.~\cite{usenix} describes their scheme to be functionality-preserving, while satisfying unremovability, unforgeability and enforcing non-trivial ownership; these properties are formally defined in their paper. We focus on the unremovability property that prevents an adversary from removing a watermark, even if s/he knows about the existence of a watermark and the used algorithm. The unremovability requires that for every PPT algorithm $\mathcal{A}$ the chance of winning the following game is negligible:
\begin{enumerate}
    \item Compute  (${M, \hat{M}}, mk, vk$) $\leftarrow{}$ MModel
    \item Run $\mathcal{A}$ and compute ${\Tilde{M}}$  $\leftarrow{}$ $\mathcal{A (O}^{f}, {\hat{M}}, vk)$\footnote{Due to the cryptographic commitment, \emph{vk} is useless for the adversary}
    \item $\mathcal{A}$ wins if: \\
    \hspace*{0.5cm}$Pr_{x \in D}[Classify (x, M) =f (x)]$ \\ \hspace*{3cm} $\approx Pr_{x \in D}[Classify (x, \Tilde{M}) =f (x)]$
    \\and $Verify (mk, vk, \Tilde{M})=0$
\end{enumerate}
We propose two computationally bounded $\mathcal{A}$'s, who not only win this security game, but also demand fewer requirements. Our black-box attack only requires oracle access to the model ($\hat{M}$) and public inputs $\Tilde{D}$ from the domain to remove the watermark while preserving the functionality. Our white-box attack does so using much less computational resources, through accessing the the model ($\hat{M}$) parameters and public inputs $\Tilde{D}$ from the domain. Although granted in the game, neither of our attacks require access to the ground truth oracle $\mathcal{O}^f$, i.e. the labeled data.

\section{Attacks on Backdoor-Based Watermarking}\label{Attacks}
The hypothesis behind our attacks is that backdoor-based watermarking schemes introduced in Section \ref{backdoor-based_WM_types} divide the input distribution into two disjoint ones: i) main distribution which is classified correctly and ii) watermark distribution which is deliberately misclassified and does not fit the main distribution. This separation in the input distribution is common among all three types of trigger sets. This results in the watermark being treated as outliers in the main classification and the network never learns to classify it \emph{correctly}. We introduce two attacks in Sections \ref{Black-Box_Attack} and \ref{White-Box_Attack} that exploit this disjointedness to remove the watermark. Our attacks remove the watermark with less requirements than the original adversary $\mathcal{A}$ \cite{usenix} (introduced in Section \ref{Security_Backdoor-based_WM}), as they do not require access to the training data and the ground truth function. They also guarantee \emph{higher efficacy}. The reason is, the unremovability game is labeled as won if the attacker $\mathcal{A}$ suggests a model $\Tilde{M}$, such that the model achieves a similar test accuracy as the watermarked model $\hat{M}$ while: $Verify (mk, vk, \Tilde{M})=0$. From the $Verify$ description in the previous section, the function outputs zero if the following holds: $\sum_{x \in T}\mathbb{I}[Classify (\Tilde{M},x) \neq T_L{x}]- \frac{1}{L}|T| > \epsilon|T|$; meaning the number of inputs in the trigger set mapped by $\Tilde{M}$ to labels \textit{different} than the pre-defined labels exceeds a negligible fraction of the trigger set. We go beyond this condition, and introduce the \emph{full removal of the watermark}, with the following two conditions:
\begin{enumerate}[label= (\roman*)]
    \item
    $Pr_{x \in D}[Classify (x, \hat{M}) =f (x)]$ \\ \hspace*{3cm} $\approx Pr_{x \in D}[Classify (x, \Tilde{M}) =f (x)]$
    \item
    $\sum_{x \in T}\mathbb{I}[Classify (\Tilde{M},x) = T_L{(x)}]- \frac{1}{L}|T| \leq \epsilon|T|$
\end{enumerate}
In full removal of the watermark, the attacker's proposed model $\Tilde{M}$, still achieves a close accuracy to the marked model's on the test set. However, in this definition, the number of inputs in the trigger set mapped by $\Tilde{M}$ to the corresponding pre-defined labels does not exceed a random labels assignment's result by more than a negligible fraction of the set. 
\subsection{Black-box Attack}\label{Black-Box_Attack}
\begin{figure}[h!]
    \centering
    \includegraphics[width=1\linewidth]{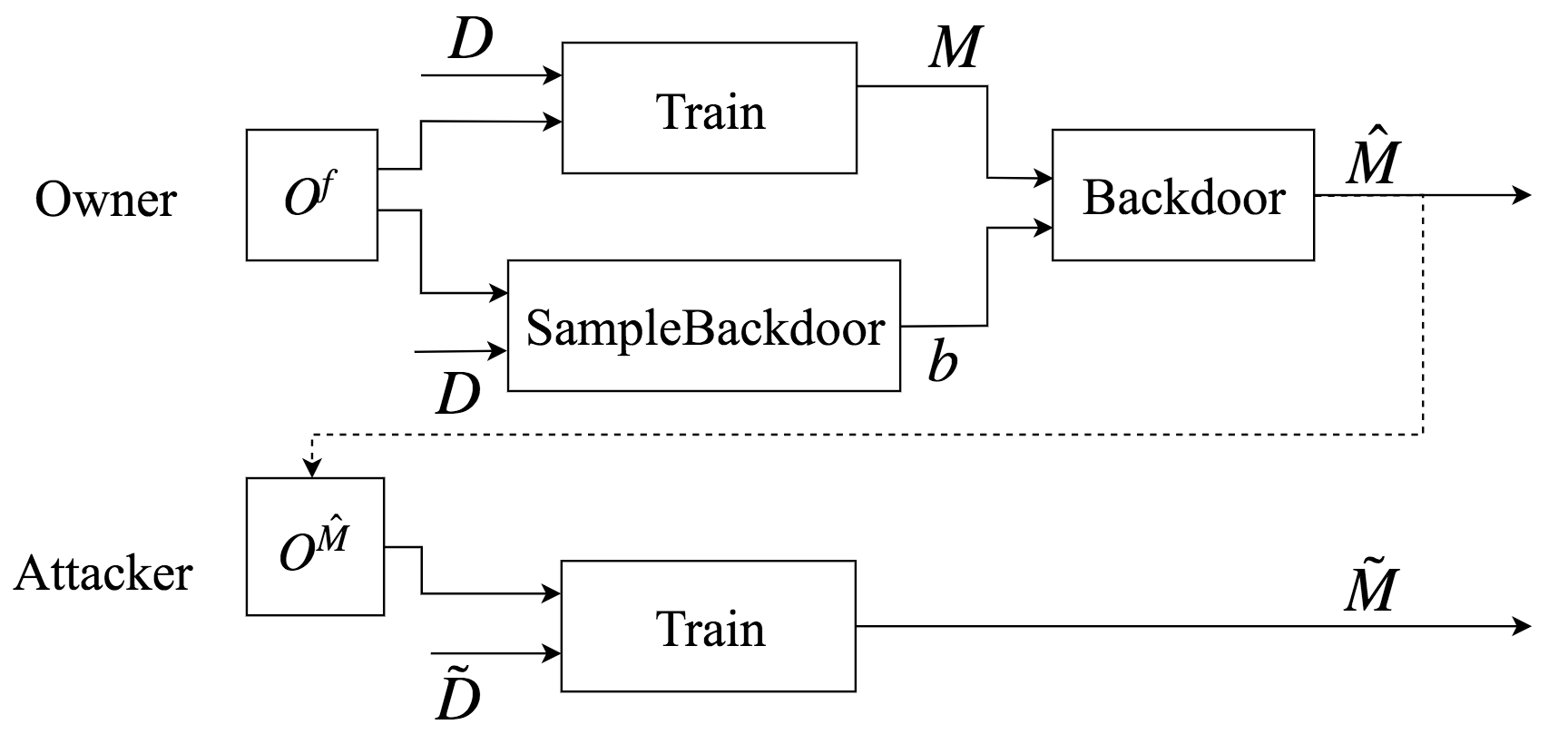}
    \caption{A schematic illustration of our black-box attack}
    \label{fig:surrogate}
\end{figure}

In our black-box attack, we steal the functionality of the marked model by querying inputs with similar distribution to the main distribution discussed earlier. Since this distribution does not include any watermarks by nature, the stolen model only copies the backdoor-free functionality. Our attack requires limited number of training inputs, and although computationally inefficient, saves on the work-intensive data preparation task. This type of model stealing is a very powerful attack and probably impossible to prevent by any watermarking scheme. Our black-box attack does not assume any access to the trigger set, the classification probability vector, any labeled data including the training data or the parameters of the watermarked model $\hat{M}$. Our attack solely relies on the public information. We query the watermarked model $\hat{M}$ with input $\Tilde{D}$ and use the classification label\footnote{Just the final class, unlike \cite{Knockoff} we do not need the probability vector} as data labels, to train a derived model as illustrated in Figure~\ref{fig:surrogate}. Note that $\Tilde{D}$ is distinct from the watermarked model $\hat{M}$'s training data $D$, but is from the same application domain. 
\\We show our attack model through the following black-box, full watermark removal game. The $\mathcal{O}^{\hat{M}}$ in the game indicates the black-box access to $\hat{M}$ through a prediction API.
\begin{enumerate}
    \item Compute  (${M, \hat{M}}, T, T_L$) $\leftarrow{}$ MModel ()
    \item Run $\mathcal{A}$ and compute ${\Tilde{M}}$  $\leftarrow{}$ $\mathcal{A (O}^{\hat{M}})$
    \item $\mathcal{A}$ wins if: \\
     (i) $Pr_{x \in D}[Classify (x, M) =f (x)]$ \\ \hspace*{3cm} $\approx Pr_{x \in D}[Classify (x, \Tilde{M}) =f (x)]$
    \\ (ii) $\sum_{x \in T}\mathbb{I}[Classify (\Tilde{M},x) = T_L{x}]- \frac{1}{L}|T| \leq \epsilon|T|$
\end{enumerate}
The attacker $\mathcal{A}$ makes queries to $\hat{M}$ and train its model $\Tilde{M}$ based on $\hat{M}$'s responses. $\mathcal{A}$ wins if it achieves the same accuracy as the original model and removes the watermark fully. 
\subsection{White-Box Attack}\label{White-Box_Attack}
The black-box attack we proposed in the previous section, does not require any information about the model parameters. However, we show that if the attacker $\mathcal{A}$ is guaranteed access to the model parameters, as is the default assumption in the unremovability game of Adi et al.'s\cite{usenix}, it can remove the watermark  more efficiently. We model the white-box attack by the following white-box full watermark removal game, which is the same as the black-box model except the oracle access to the model  $\mathcal{O}^{\hat{M}}$ is replaced by a direct access to $\hat{M}$.
\begin{enumerate}
    \item Compute  (${M, \hat{M}}, T, T_L$) $\leftarrow{}$ MModel ()
    \item Run $\mathcal{A}$ and compute ${\Tilde{M}}$  $\leftarrow{}$ $\mathcal{A} (\hat{M})$
    \item $\mathcal{A}$ wins if: \\
     (i) $Pr_{x \in D}[Classify (x, M) =f (x)]$ \\ \hspace*{3cm} $\approx Pr_{x \in D}[Classify (x, \Tilde{M}) =f (x)]$
    \\ (ii) $\sum_{x \in T}\mathbb{I}[Classify (\Tilde{M},x) = T_L{x}]- \frac{1}{L}|T| \leq \epsilon|T|$
\end{enumerate}
Recall that our goal is to prevent the model from learning the misclassifications which are essential to watermarking. As mentioned earlier, we see the watermarking samples as outliers to the main distribution, and believe that the marked model overfits to learn misclassifying them. In order to remove the watermark, we apply regularization to the marked model to \emph{normalize} the weights and avoid overfitting~\cite{ML_Ref}.
\begin{figure}[h!]
    \centering
    \includegraphics[width=1\linewidth]{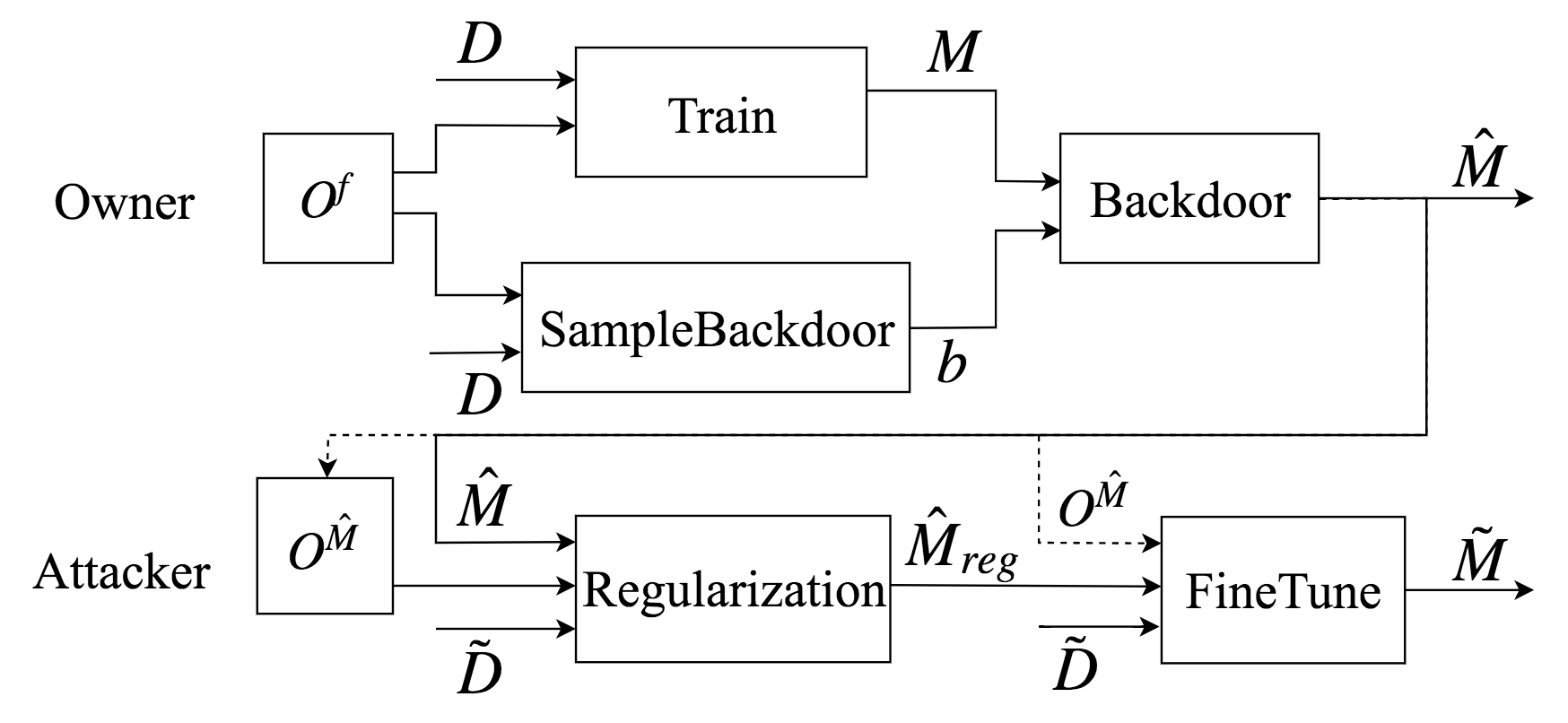}
    \caption{A schematic illustration of our white-box attack}
    \label{fig:WBattack}
\end{figure}
Our white-box attack, illustrated in Figure~\ref{fig:WBattack}, consists of the two following sub algorithms: regularization and fine-tuning. The input for both sub-algorithms is $\Tilde{D}$, which is from the same domain as but distinct from $D$. 
\begin{enumerate}
    \item $\mathcal{A}_{Reg} (\hat{M}, \mathcal{O}^{\hat{M}})$ $\rightarrow$ $\hat{M}_{Reg}$
    \item $\mathcal{A}_{Fine} (\hat{M}_{Reg}, \mathcal{O}^{\hat{M}})$ $\rightarrow$ $\Tilde{M}$
\end{enumerate}
The first sub-algorithm $\mathcal{A}_{Reg}$ performs the regularization on $\hat{M}$. Since we do not know which layer contributes to learning the watermark misclassification, we define regularization to impact all layers to prevent overfitting to the backdoors.
Our experiments show that $\mathcal{A}_{Reg}$ removes the watermark fully by using L2 regularization. However, it affects the test accuracy compared to the original model $M$. To compensate for this accuracy reduction, the output of $\mathcal{A}_{Reg}$ is then fed to $\mathcal{A}_{Fine}$ to be fine-tuned on an unmarked training set. 
We emphasize  that our white-box attack does not require any information of the ground truth function or the trigger set for winning the game. Instead, it uses a random set of inputs from the domain of the original model and queries the model $\hat{M}$ to label them. Our experiments show that this attack is significantly more efficient than training a new model and achieves the same accuracy.
\subsection{Property Inference Attack}\label{PropertyInference_Attack}
\begin{figure}[h!]
    \centering
    \includegraphics[width=\linewidth]{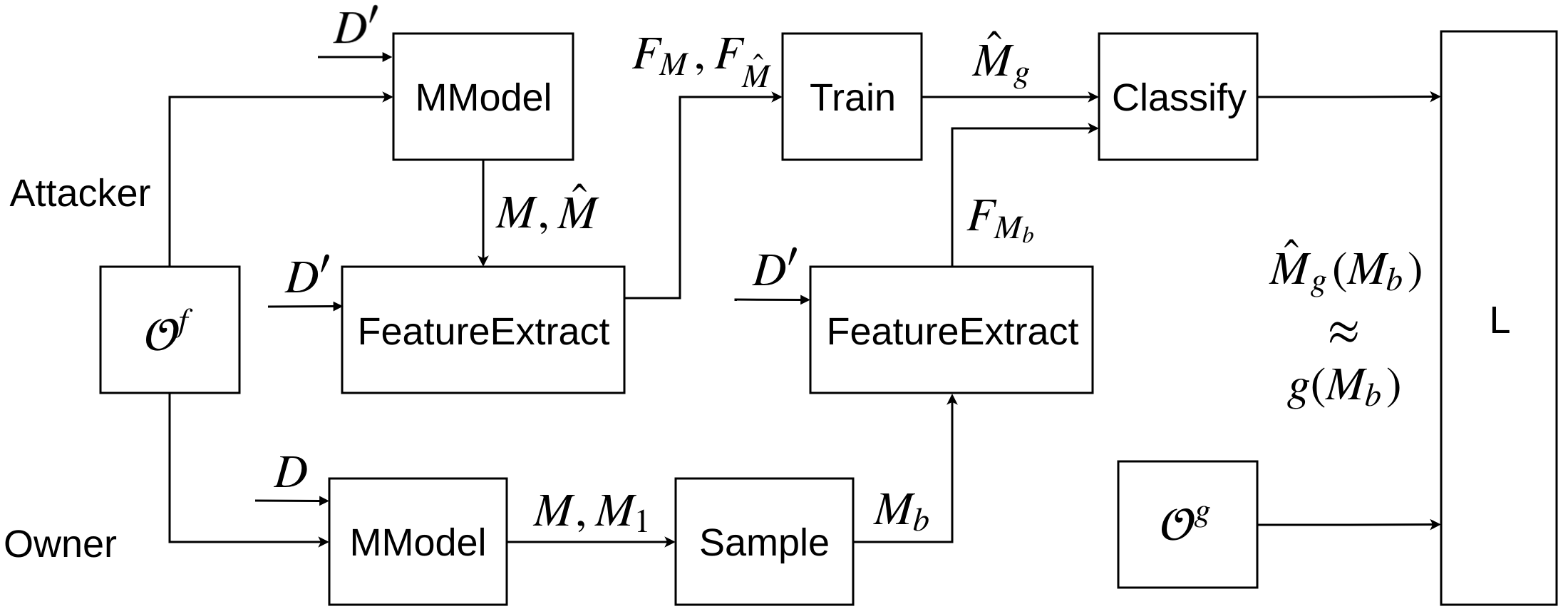}
    \caption{Illustration of our property inference attack}
    \label{fig:property_inference}
\end{figure}

Property inference attacks \cite{ganju2018property} have been originally proposed to extract knowledge about training data given white-box access to a neural network. We propose to use a property inference attack to detect whether a backdoor-based watermark has been embedded in a neural network. In case an embedded watermark is detected, a removal attack is applied. The property inference attack could also verify the success of the model stealing attack in removing the embedded watermark. In our attack, the attacker needs to have access to some oracle $\mathcal{O}^f$, the backdoor-based watermark embedding algorithm $MModel$ and has to be able to generate $k$ sufficiently different high-accuracy models for $f$. Note that $f$ does not need to be the same as the function in the attacked model. In the following part we present the watermark detection security game. 
Assume $g$ is a binary classifier that labels an input model as marked or non-marked. Given a model $\hat{M}$ and an oracle access to the function $g$, $\mathcal{O}^{g}$, the attacker wins if he can design a classifier $\hat{M}_g$ that agrees for the classification of a given model with $\mathcal{O}^{g}$ with a probability of at least $1-\epsilon$. Figure \ref{fig:property_inference} illustrates this watermark detection game.
Formally, $\mathcal{A}$ wins the watermark detection game as following. 
\begin{enumerate}
    \item Compute $(M_0, M_1, mk, vk) \leftarrow MModel$() 
    \item Sample $b \xleftarrow[]{\$} \{0,1\}$
     \item Run $\mathcal{A}$ to compute $\hat{M}_g \leftarrow \mathcal{A}(\mathcal{O}^f)$
    \item $\mathcal{A}$ wins if $Pr[Classify(M_b, \hat{M}_g) \neq g(M_b)] \leq \epsilon$
\end{enumerate}
The cornerstone to the attack is the property inference algorithm which extracts a set of labeled feature vectors from a non-watermarked model $M$ and its watermarked counterpart $\hat{M}$. The binary label denotes whether the feature vector was extracted from $M$ or $\hat{M}$. Given sufficiently many training examples, the intuition is to generate a feature space that is clearly separable between the two classes. 
Embedding watermark in a model makes a meaningful difference in model parameters. In property inference attack we try to capture features from model parameters that demonstrate this difference. 
This method is referred to as $FeatureExtract(M, x)$ where $M$ is the input model and $x$ is part of the benign training data. Next, we introduce the function $PIData$ which generates the training data by extracting features from the non-watermarked $M$ and its watermarked counterpart $\hat{M}$. 
\\
$PIData$(): 
\begin{enumerate}
    \item Generate $(M, \hat{M}, mk, vk) \leftarrow MModel$()
    \item Extract $F_M = FeatureExtract(M, x)$
    \item Extract $F_{\hat{M}} = FeatureExtract(\hat{M}, x)$
    \item Output $F_M , F_{\hat{M}}$
\end{enumerate}
Each invocation of $PIData$ requires that the attacker generates a new high-accuracy model on some task $f$. The attack $PIAttack$ generates training and testing data and stops training the binary classifier $\hat{M}_g$ once the testing accuracy is sufficiently high.
\\
$PIAttack$():
\begin{enumerate}
    \item Generate $(D_{train}, D_{test}) \leftarrow \bigcup\limits_{i=1..k} PIData()$
    \item Compute $\hat{M}_g \leftarrow Train(D_{train})$
    \item Stop when $Pr_{F \in D_{test}}[Classify(F, \hat{M}_g) \neq g(F)] \leq \epsilon$
    \item Output $\hat{M}_g$
\end{enumerate}
In our property inference attack, the adversary is granted white-box access to the target marked model. This implies an approximation of the function $f$ and consequently an approximation of the dataset which we use in order to select a somewhat similar dataset in $PIData$. In order to design a powerful $PIData$ feature extraction algorithm, we investigated the effectiveness of a variety of model features in watermark detection and ultimately selected the activations in the last convolution layer as the most effective features. We form sets out of these features and leverage the DeepSets \cite{DeepSets} architecture to develop the binary classifier $\hat{M}_g$ over the sets to detect the presence of a watermark in the model. This approach is inspired by Ganju et al's work \cite{ganju2018property} on property inference attack in fully connected neural networks. In Section \ref{Experiments}, we demonstrate our property inference attack detects the presence of watermark in the target model. 

\section{Experiments}\label{Experiments}
We present the results of applying our black-box and white-box attack on backdoor-based watermarking scheme of Section \ref{backdoor-based_WM_types}; these results confirm  that our attacks successfully remove the watermarks. We proceed by presenting the results for our property inference attack. 
\subsection{Experiment Setup}\label{Experiment_Setup}
In Section \ref{Attacks}, we introduced our attacks through full watermark removal games between a challenger $MModel$ and the attacker $\mathcal{A}$. We simulate both entities in this section and run experiments according to the described algorithms. 

\subsubsection{Data Sets and Models} We evaluate our attacks over three popular data sets in DNN literature: MNIST, CIFAR-10 and CIFAR-100. 
For MNIST dataset, we use LeNet model and train it on 60K training images and test it on 10K test images. For CIFAR-10, we use VGG-16 and train the model on 50K training images and test it on 10K test images. For MNIST and CIFAR-10 datasets we split the training data in half for the attacker and owner. Our mini-batches contain 64 elements and we use the \emph{RMSProp} \cite{ML_Ref} optimizer with learning rate of $0.001$. While training any model, we use \emph{Early Stopping}~\cite{ML_Ref} on the training accuracy with a min-delta of 0.1\% and a patience of 2. Unlike the other two datasets, for CIFAR-100 we use ResNet-32 and train it on 50K training images and test it on 10K test images. For CIFAR-100 we use overlapping training data for the attacker and owner, we discuss the reasons in Section \ref{Overlap}. We also use the batch size of 100, SGD optimizer with initial learning rate = 0.1 and momentum = 0.9. We adjust the learning rate by dividing it by 10 every 30 epochs. For the white-box attack we use early stopping on the watermark retention with a baseline of 0.1 a patience of 2 for Embedded Content and Pre-specified Noise watermarking schemes and 0 for Abstract Images; these watermarking schemes were introduced in Section \ref{backdoor-based_WM_types}. 
We use 0 - 1 normalization \cite{ML_Ref} for all datasets.

\subsubsection{Original and Watermarked Model Generation} 
\begin{figure}[h!]
    \centering
    \includegraphics[width=1\linewidth]{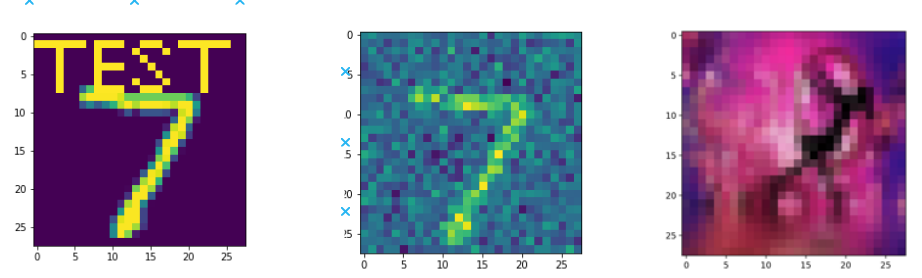}
    \caption{Samples of watermarks embedded by \emph{MModel}\\  (a)Content Embedded  (b)Gaussian Noise   (c)Abstract Images}
    \label{fig:Wsamples}
\end{figure}

We first simulate the $MModel$ algorithm in our full watermark removal games to generate the original model $M$, the watermarked model $\hat{M}$, and the watermark consisting of the watermark set $T$ and its corresponding labels $T_L$. The watermark set for different schemes, shown through examples in Figure \ref{fig:Wsamples}, is constructed according to Section \ref{backdoor-based_WM_types}. 
$MModel$ trains a model $\hat{M}$ with a portion the of the watermark set and a portion of the remaining training set. Note that the rest of the two sets is needed to form \textit{test set} and \textit{watermark test set}. Recall from Section \ref{backdoor-based_WM_types} that the watermark set and the watermark test set are the same for the Abstract Images watermarking scheme, but not for the Embedded Content or the Pre-spesified Noise scheme. 

\subsubsection{Attack Algorithm $\mathbf{\mathcal{A}}$ and Generating $\mathbf{\Tilde{M}}$} 
In both our black-box and white-box attacks, the algorithm $\mathcal{A}$ aims to derive a model $\Tilde{M}$ that keeps the same test accuracy as the original model $M$, while it reduces the watermark retention to  $\frac{1}{|L|}$, where $|L|$ is the total number of valid classes. This reduction, shows that the model associates the watermarked input to the pre-defined class, not more than a random classifier would do, hence indicating success in removing the watermark fully. To generate $\Tilde{M}$, neither of our attacks use the original model $M$'s training data with the ground truth labels, nor any of the watermarking information. Instead, they both query the watermark model $\hat{M}$ with inputs from the publicly known domain of $\hat{M}$ and train $\Tilde{M}$ with the corresponding labels. The white-box attack initiates the $\Tilde{M}$ with the parameters of $\hat{M}$, then undergoes a regularization followed by fine-tuning with public data labeled by $\hat{M}$.

\subsubsection{Attack Algorithm $\mathbf{PIAttack}$ and Generating $\mathbf{\hat{M}_g}$} \label{PIAttack}
As described in Section \ref{PropertyInference_Attack}, our $PIAttack$ algorithm trains a binary classifier over the model features extracted by $PIData$. In our experiments $PIAttack$ is trained on a combination (20) of (10) marked and (10) unmarked models for a similar dataset as the target model's. In the following procedure -- using CIFAR-10 as an example dataset -- we show how $PIData$ operates on each of these models in the combination to extract features for $PIAttack$:  i) The model is fed with 10,000 test images, ii) The 10,000 output sets from the activations in the last convolution layer of the model is collected. iii) These output sets are divided into different groups according to their corresponding label, e.g.~ten groups are formed in this step for CIFAR-10 with approximately 1000 sets in each group iv) A bootstrapping round is defined as sampling 100 outputs with replacement from the group. Ten rounds of bootstrapping are applied on all ten groups, these ten rounds are combined to obtain 100 sets of samples from each group. 
We apply this procedure to all 20 models in the combination and obtain 2,000 sets of samples. We use a binary classifier based on DeepSets \cite{DeepSets}. Our classifier contains of two parts: i) an encoder and ii) a classifier. The encoder consists of four fully-connected ReLU dense layers with 300, 200, 100 and 30 neurons each. The encoder is applied to every filter of the input, the results are added up after encoding and are fed to the classifier, which consists of three fully-connected ReLU dense layers of 512 neurons each and a one-neuron dense layer with sigmoid activation function as the output layer. We use Adam optimizer, learning rate of 0.001, and binary cross-entropy as loss function. We train the model for 100 epochs.

\subsubsection{Security and Performance Evaluation} In what follows, we present our black-box and white-box attack with concrete parameters in their setup and evaluation. As mentioned earlier in this section, our security evaluation metrics are: i) test accuracy and ii) watermark retention.
For test accuracy, we compare the accuracy of model generated by our attack with that of the target model on classifying an unseen test set. For watermark retention, we measure what portion of a set of marked inputs are classified as their pre-defined labels by the models generated through our attacks. We also evaluate the performance of our attacks, based on the time they take to run rather than the number of epochs. As the number of epochs depends on some factors in the model training such as input size, while the time is an independent measure. For example an epoch in the fine-tuning phase (Section \ref{White-Box_Attack}) takes much more time than an epoch in the regularization, as the size of the training set in fine-tuning is at least ten times the training set size in the regularization.

\begin{figure*}
\subfloat[Black-box Attack, Content, MNIST]{\includegraphics[width=.33\textwidth]{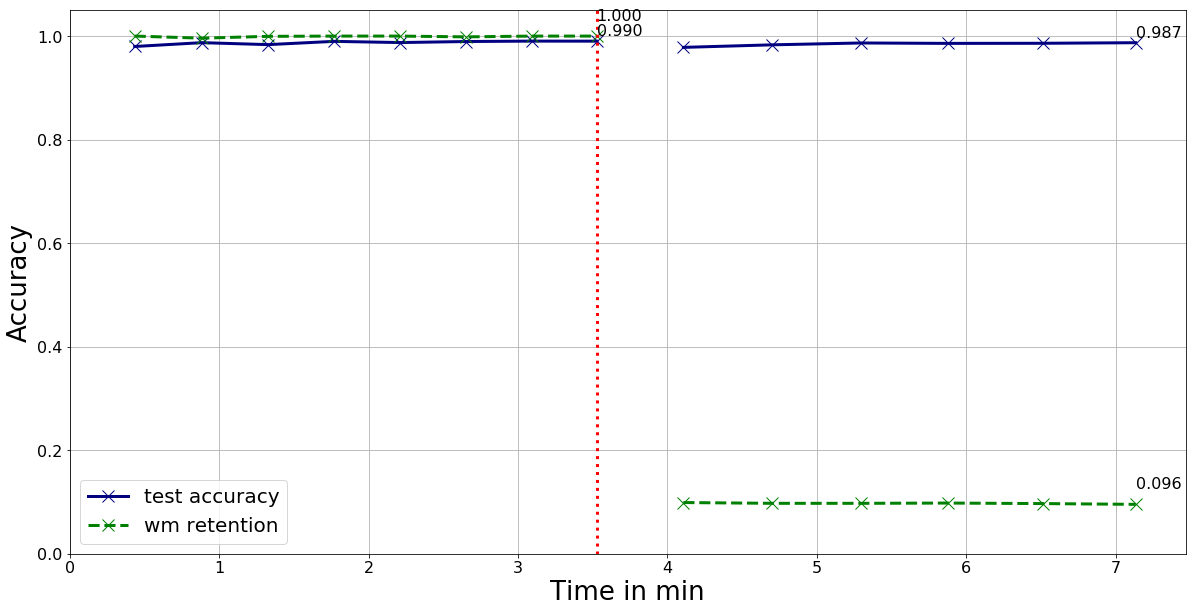}} 
\subfloat[Black-box Attack, Noise, MNIST]{\includegraphics[width=.33\textwidth]{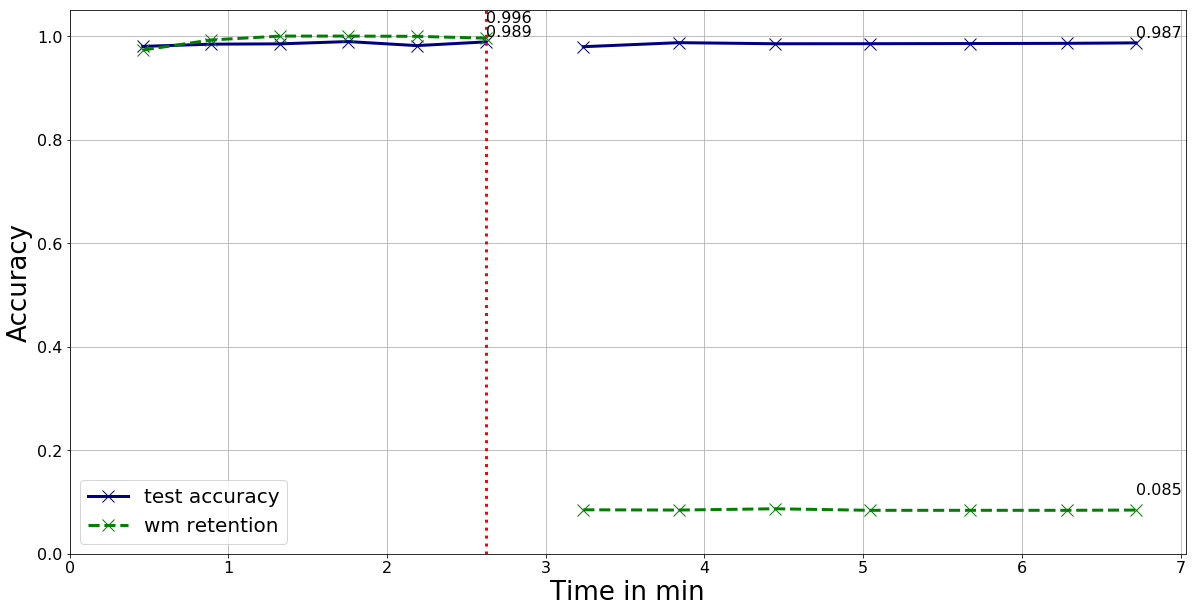}} 
\subfloat[Black-box Attack, Abstract, MNIST]{\includegraphics[width=.33\textwidth]{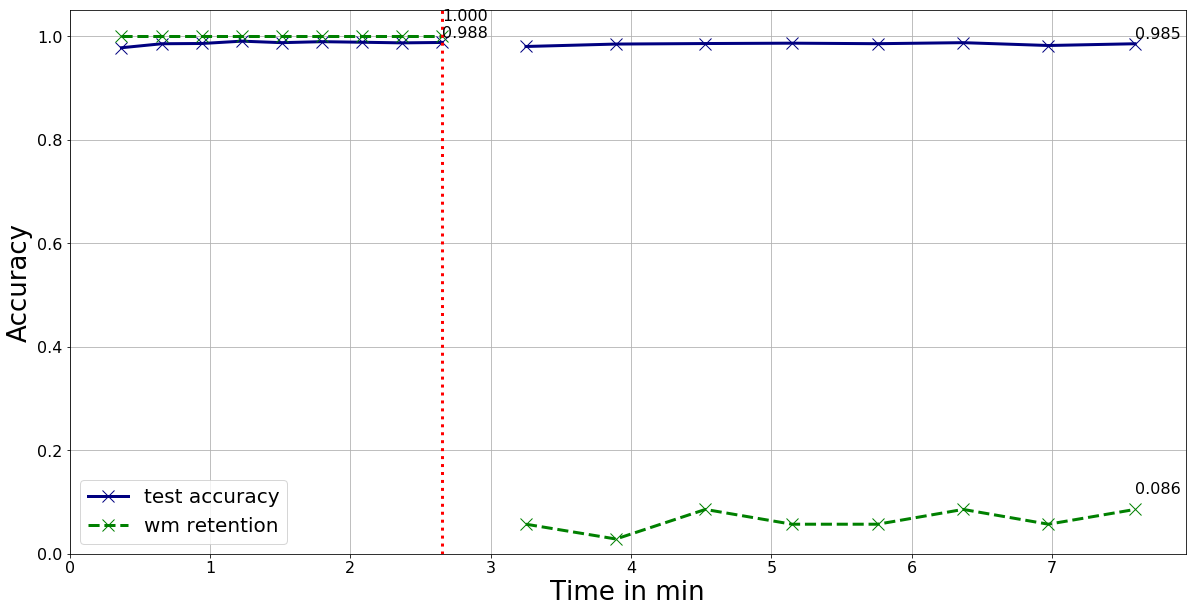}} 
\hfill
\subfloat[White-box Attack, Content, MNIST]{\includegraphics[width=.33\textwidth]{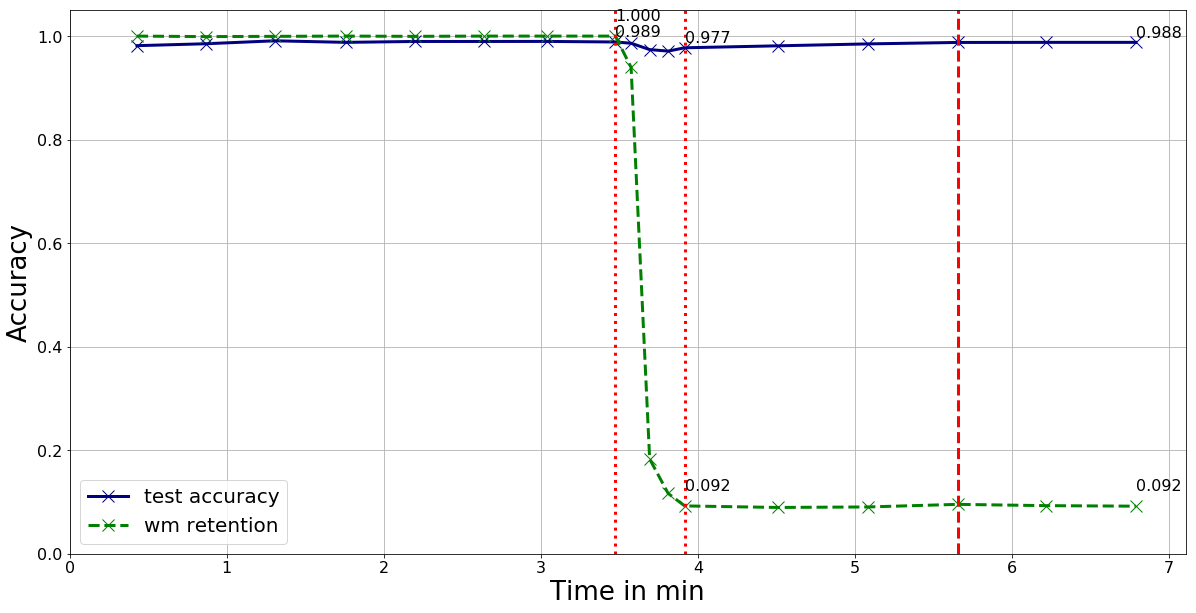}}
\subfloat[White-box Attack, Noise, MNIST]{\includegraphics[width=.33\textwidth]{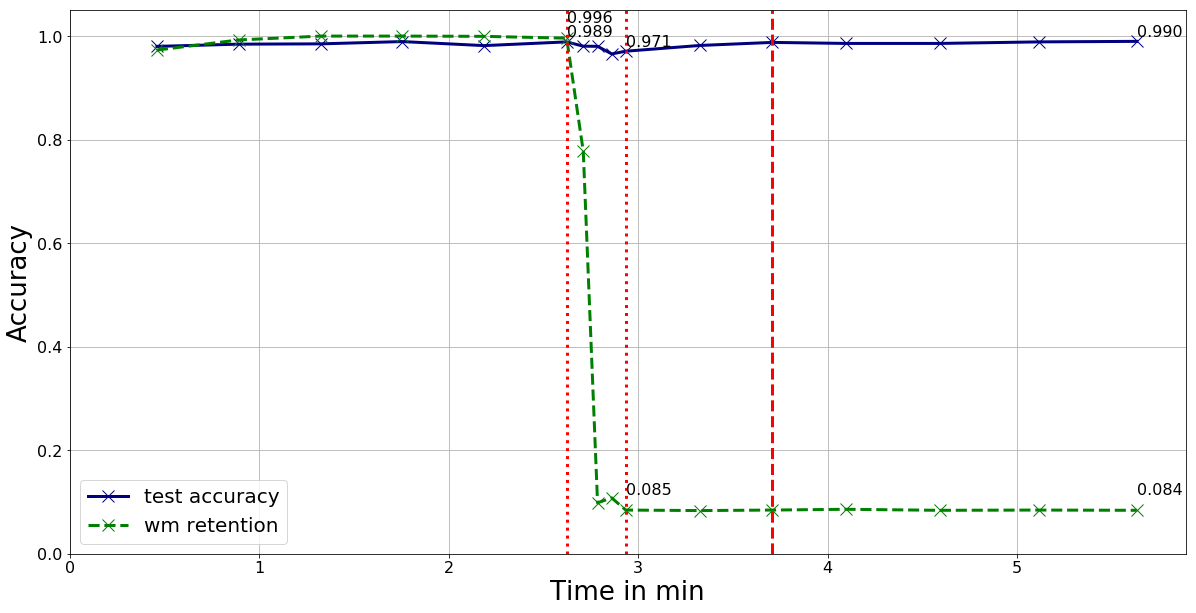}}
\subfloat[White-box Attack, Abstract, MNIST]{\includegraphics[width=.33\textwidth]{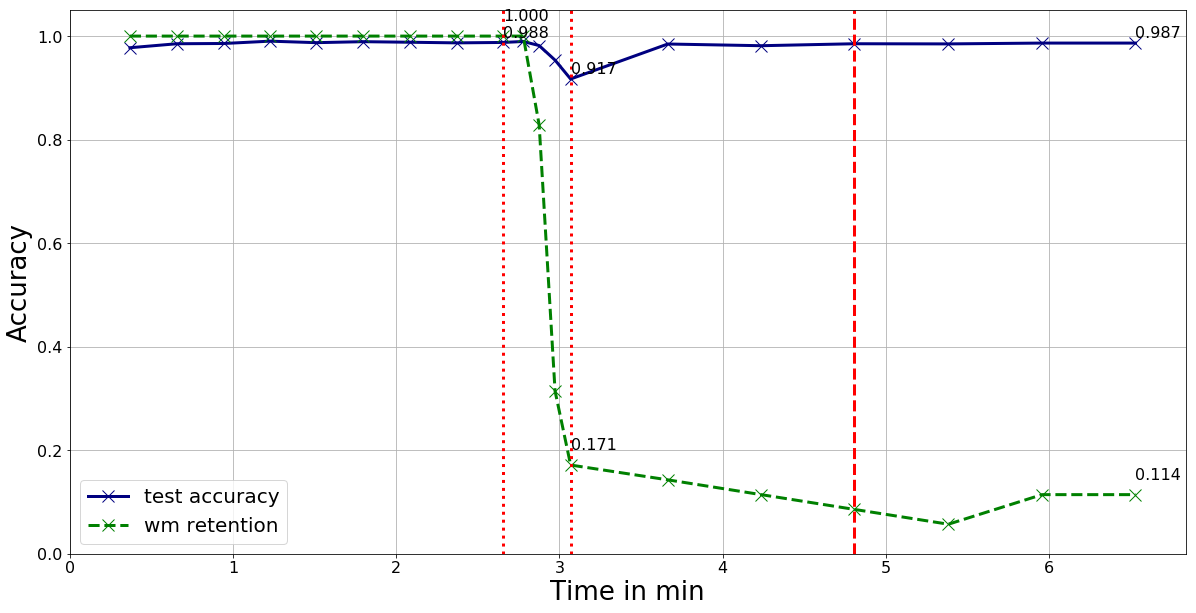}}\hfill
\caption{Experiment results on MNIST for various backdoor-based watermarking schemes: (a,b,c) represent black-box attack results and (d, e,f) demonstrate the white-box attack results, respectively on the watermarks: Embedded Content (a,d), Pre-specified Noise (b,e), and Abstract Images (c,f)}
\label{fig:Res_MNIST} 
\end{figure*}

\begin{figure*}
\subfloat[Black-box Attack, Content, CIFAR-10]{\includegraphics[width=.33\textwidth]{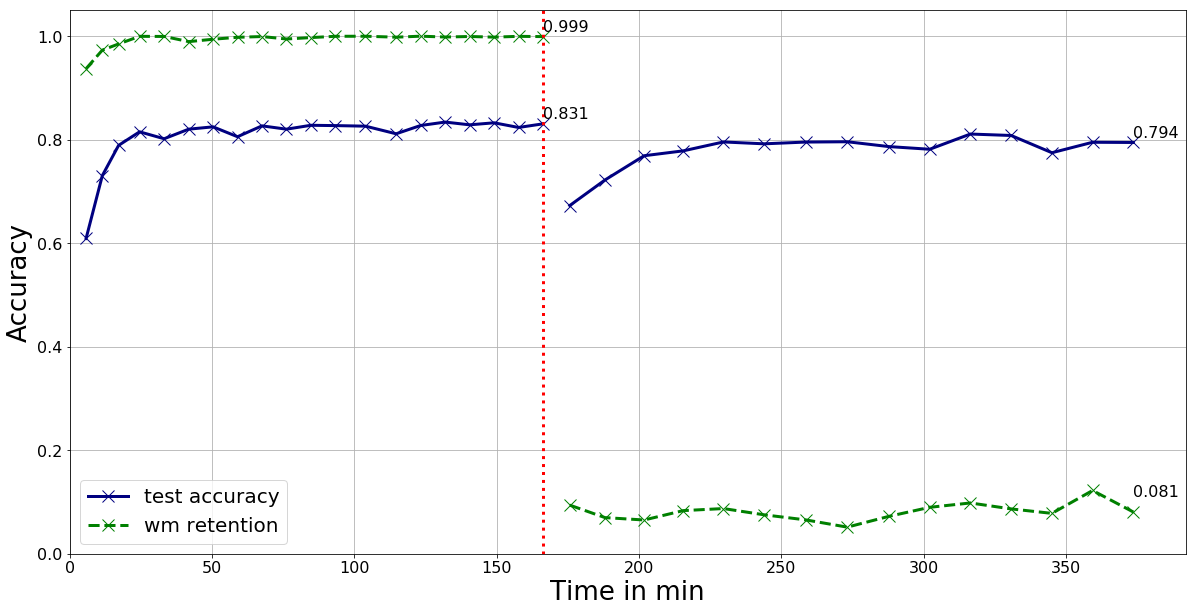}} 
\subfloat[Black-box Attack, Noise, CIFAR-10]{\includegraphics[width=.33\textwidth]{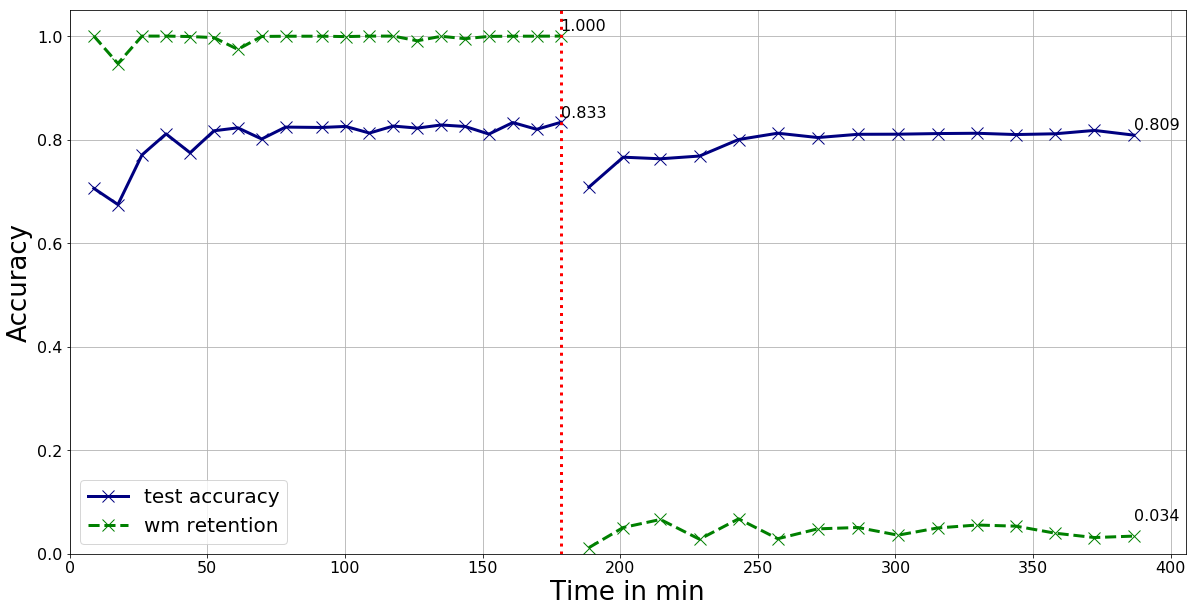}}  
\subfloat[Black-box Attack, Abstract, CIFAR-10]{\includegraphics[width=.33\textwidth]{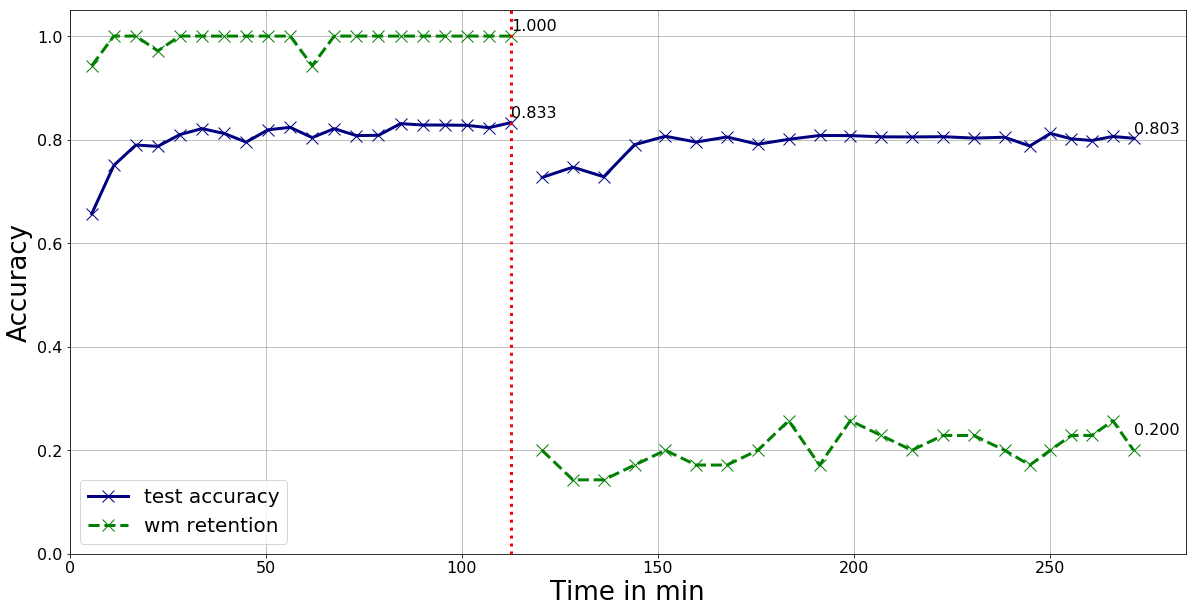}}\hfill
\subfloat[White-box Attack, Content, CIFAR-10]{\includegraphics[width=.33\textwidth]{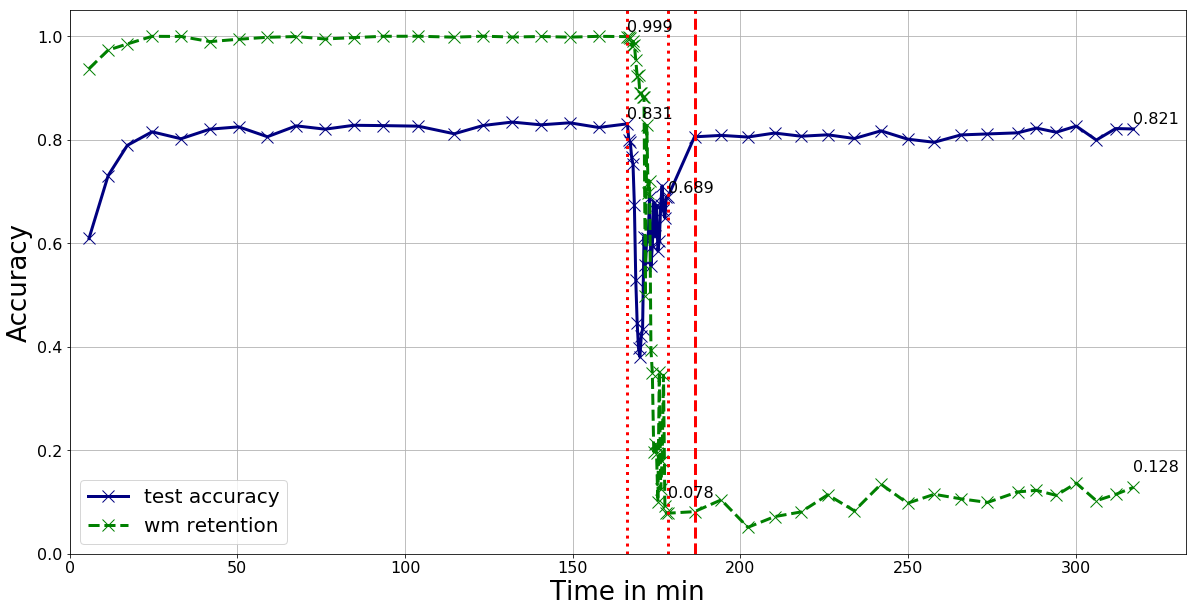}}
\subfloat[White-box Attack, Noise, CIFAR-10]{\includegraphics[width=.33\textwidth]{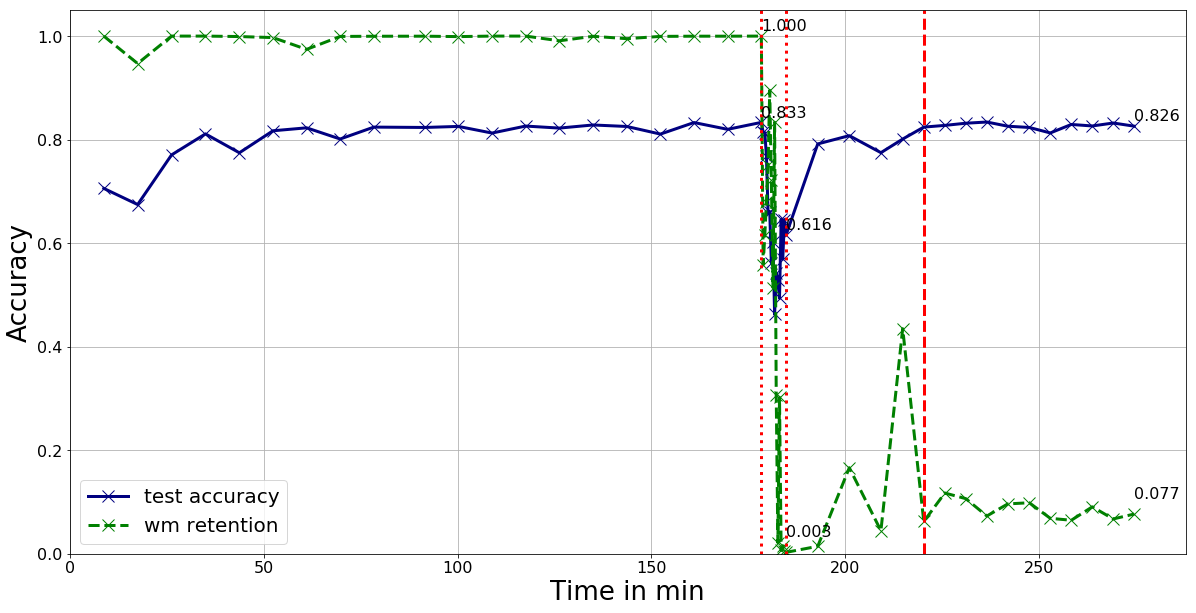}}
\subfloat[White-box Attack, Abstract, CIFAR-10]{\includegraphics[width=.33\textwidth]{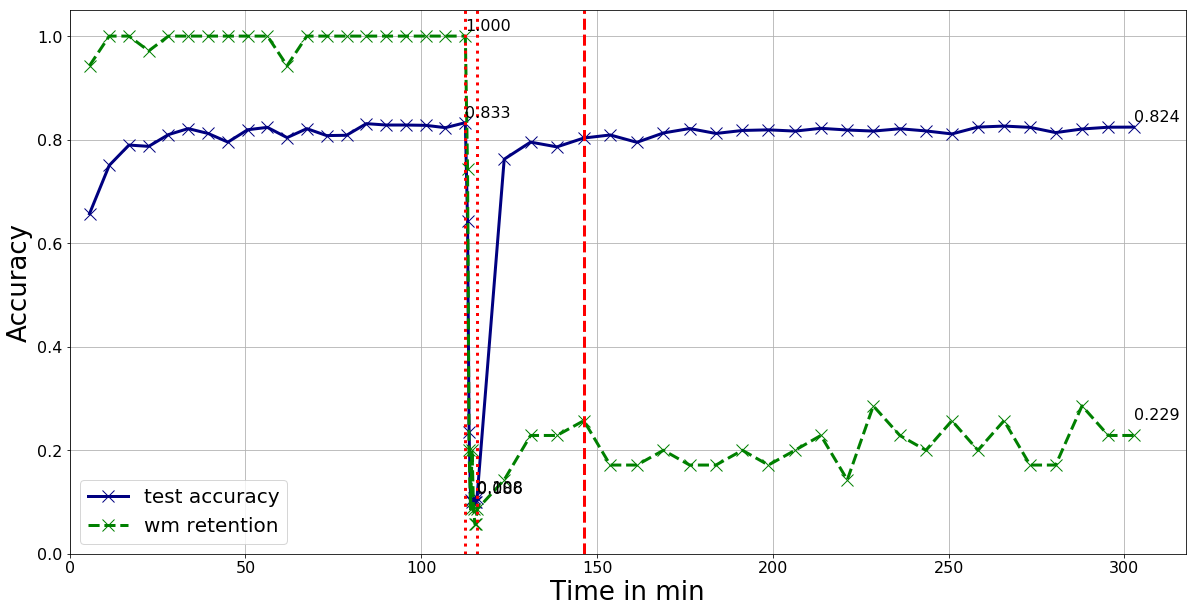}}\hfill
\caption{Experiment results on CIFAR-10 for various backdoor-based watermarking schemes: (a,b,c) represent black-box attack results and (d, e,f) demonstrate the white-box attack results, respectively on the watermarks: Embedded Content (a,d), Pre-specified Noise (b,e), and Abstract Images (c,f)} 
\label{fig:Res_CIFAR} 
\end{figure*}


\begin{figure*}
\centering
\subfloat[Black-box Attack, Abstract, CIFAR-100]{\includegraphics[width=.33\textwidth]{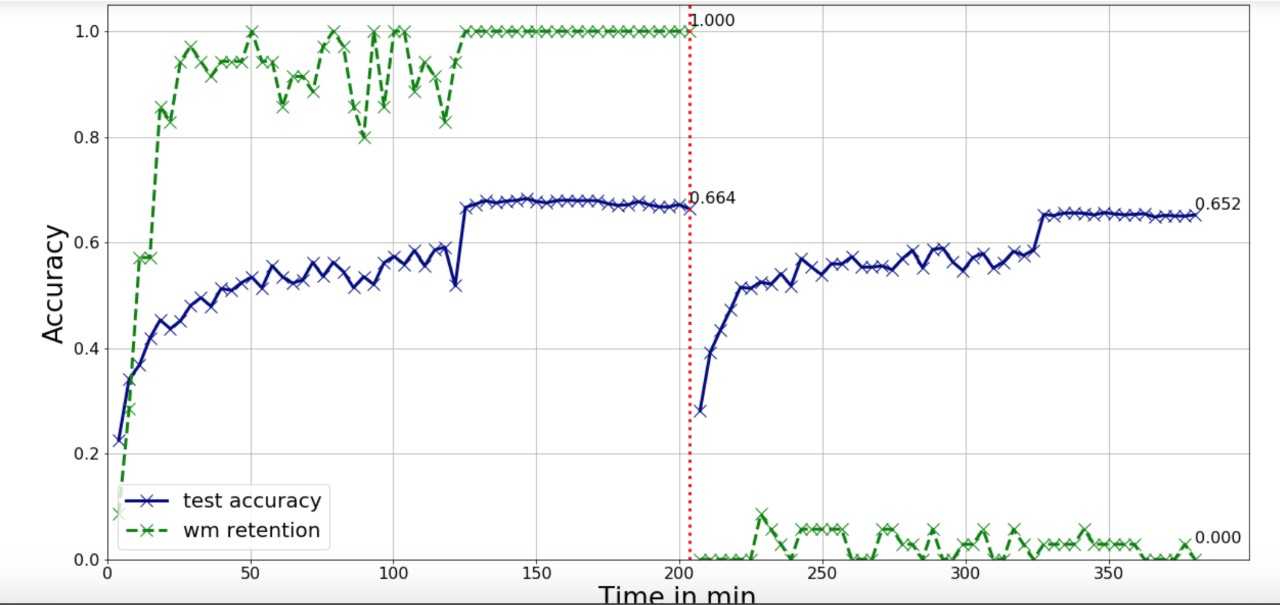}}
\subfloat[White-box Attack, Abstract, CIFAR-100]{\includegraphics[width=.33\textwidth]{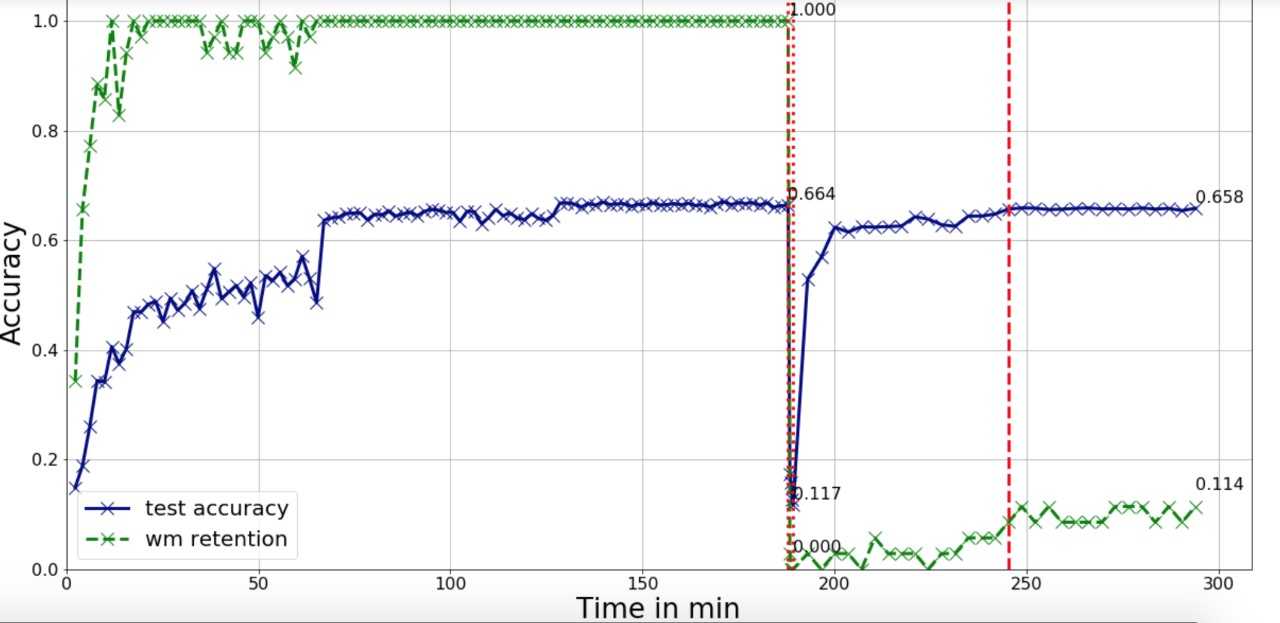}}\hfill
\caption{Experiment results on CIFAR-100, (a) black-box attack and (b) white-box attack on Abstract Images watermark}
\label{fig:Res_CIFAR_100} 
\end{figure*}

\subsection{Results}\label{Results}
\subsubsection{Watermark Removal}
We present the results of our black-box and white-box attacks. We evaluate our attacks on MNIST and CIFAR-10 datasets in Figures \ref{fig:Res_MNIST} - \ref{fig:Res_CIFAR}, over each of the three watermarking schemes described in Section \ref{backdoor-based_WM_types}, namely: Embedded Content $ (a, d)$, Pre-specified Noise $ (b, e)$ and Abstract Images $ (c, f)$. As the Abstract Images in its original paper \cite{usenix} is used for watermarking models trained on CIFAR-100 too, we include experiments for watermark removal for those models as shown in Figure \ref{fig:Res_CIFAR_100}. The subfigures $ (a)$, $ (b)$ and $ (c)$ in Figures \ref{fig:Res_MNIST} - \ref{fig:Res_CIFAR}, and Figure \ref{fig:Res_CIFAR_100} (a), depict the black-box attack results. Similarly, the subfigures $ (d)$, $ (e)$ and $ (f)$ in Figures \ref{fig:Res_MNIST} - \ref{fig:Res_CIFAR}, and Figure \ref{fig:Res_CIFAR_100} (b), indicate the white-box attack results on the corresponding data set. Each graph evaluates the models generated by algorithm $MModel$ followed by $\mathcal{A}$'s evaluations, where the former represents the owner's watermarked model and the latter represents the attacker's model. As shown in black-box attack graphs, $ (a)$, $ (b)$ and $ (c)$, the attacker $\mathcal{A}$ starts training its model $\Tilde{M}$, when $MModel$ is done training the watermark model $\hat{M}$ (red dotted line). $\mathcal{A}$ initiates training from random weights and queries $\hat{M}$ for labeling its input to train the model afterwards. The graphs indicate how long the black-box attack takes to train $\Tilde{M}$ compared to the time $MModel$ needs to spend to train $\hat{M}$. In both models, the training continues until its test accuracy is stable at a desired level. For the white-box attack on the other hand, subfigures $ (d)$, $ (e)$ and $ (f)$, $\mathcal{A}$ initiates the algorithm from $MModel$ parameters (first red dotted line) in addition to requiring $\hat{M}$ labeling its inputs. $\mathcal{A}$ first goes through a regularization phase for $\Tilde{M}$ that takes a short time compared to the model's training time. After removing the watermark (second red dotted line), $\mathcal{A}$ applies fine-tuning until $\Tilde{M}$ is $\epsilon$-close to the $\hat{M}$ in test accuracy (red dashed line). The results of our experiments are summarized in Tables \ref{tab:BB-MNIST}- \ref{tab:PERF-CIFAR100}. The watermarks in the table correspond to Embedded Content, Pre-specified Noise, and Abstract Images schemes as introduced in Section \ref{backdoor-based_WM_types}.
\begin{table}[h!]
  \centering 
  \caption{Black-box attack on MNIST}
  \label{tab:BB-MNIST}
  \begin{tabular}{ccc}
    \toprule
    Watermark & Acc. Drop & Watermark Ret. \\
    \midrule
    Content & 0.3\% & 9.6\% \\
    Noise & 0.2\% & 8.5\% \\
    Abstract & 0.2\% & 8.6\% \\
    \bottomrule
  \end{tabular}
\end{table}
\begin{table}[h!]
  \centering
  \caption{White-box \& Black-box attacks on MNIST}
  \label{tab:PERF-MNIST}
  \begin{tabular}{cccc}
    \toprule
    Watermark & WM Model & Black-box & White-box\\
    \midrule
    Content & 3.5m & 3.6m & 0.59m\\
    Noise & 2.6m & 4.1m & 0.77m\\
    Abstract  & 2.6m & 5m & 2.1m\\
    \bottomrule
  \end{tabular}
\end{table}
\begin{table}[h!]
  \centering
  \caption{Black-box attack on CIFAR-10}
  \label{tab:BB-CIFAR10}
  \begin{tabular}{ccc}
    \toprule
    Watermark & Acc. Drop & Watermark Ret. \\
    \midrule
    Content & 3.7\% & 8.1\% \\
    Noise & 2.4\% & 3.4\% \\
    Abstract & 3\% & 20\% \\
    \bottomrule
  \end{tabular}
\end{table}
\begin{table}[h!]
  \centering
  \caption{White-box \& Black-box attacks on CIFAR-10}
  \label{tab:PERF-CIFAR10}
  \begin{tabular}{cccc}
    \toprule
    Watermark & WM Model & Black-box & White-box\\
    \midrule
    Content & 166m & 244m & 20m\\
    Noise & 178m & 208m & 20m\\
    Abstract  & 112m & 158m & 33m\\
    \bottomrule
  \end{tabular}
\end{table}
\begin{table}[h!]
  \centering
  \caption{Black-box attack on CIFAR-100}
  \label{tab:BB-CIFAR100}
  \begin{tabular}{ccc}
    \toprule
    Watermark & Acc. Drop & Watermark Ret. \\
    \midrule
    Abstract & 1.2\% & 1\% \\
    \bottomrule
  \end{tabular}
\end{table}
\begin{table}[h!]
  \centering
  \caption{White-box \& Black-box attacks on CIFAR-100}
  \label{tab:PERF-CIFAR100}
  \begin{tabular}{cccc}
    \toprule
    Watermark & WM Model & Black-box & White-box\\
    \midrule
    Abstract & 203m & 177m & 10m\\
    \bottomrule
  \end{tabular}
\end{table}
Note that in the performance comparison in Tables \ref{tab:PERF-MNIST}, \ref{tab:PERF-CIFAR10} and \ref{tab:PERF-CIFAR100}, the white-box and the black-box attacks have the same accuracy. 
Our results show that the black-box attack removes the watermark with a performance comparable to that of the watermarked model, whereas the white-box attack does so with significant speed-up. It is also worth noting that by allowing the fine-tuning to continue for longer, the white-box attack can even achieve higher accuracy than the black-box attack.

\subsubsection{Property Inference Attack}
We investigated the effectiveness of our property inference attack to distinguish watermarked models from unmarked ones as shown in Table \ref{tab:PI_CIFAR}. In our experiment, we investigate whether we can detect the presence of Abstract Images as embedded watermark into a target model. 
\begin{table*}[t]
  \centering
  \caption{Property Inference Attack on models trained/tested on CIFAR-10 or CIFAR-100, ``10" stands for CIFAR-10 and ``100 (i)" refers to the data corresponding to the $i^{th}$ 10 classes of CIFAR-100}
  \label{tab:PI_CIFAR}
  \begin{tabular}{cccccccccccc}
    \toprule
    \backslashbox{Test}{Train} & 10 & 100 (1) & 100 (2) & 100 (3) & 100 (4) & 100 (5) & 100 (6) & 100 (7) & 100 (8) & 100 (9) & 100 (10) \\
    \midrule
    10      & -- & 63\% & 74\% & 74\% &  68\% & 82\% & 81\% & 85\% & 77\% & 74\% & 60\% \\
    100 (1) & 76\% & -- & 80\% & 75\% & 67\% & 66\% & 81\% & 76\% & 70\% & 72\% & 72\% \\
    100 (2) & 75\% & 66\% & -- & 80\% & 70\% & 75\% & 83\% & 76\% & 70\% & 69\% & 67\% \\
    100 (3) & 82\% & 67\% & 78\% & -- & 68\% & 78\% & 82\% & 83\% & 72\% & 73\% & 65\% \\
    100 (4) & 74\% & 64\% & 75\% & 76\% & -- & 74\% & 80\% & 79\% & 71\% & 71\% & 62\% \\
    100 (5) & 86\% & 67\% & 78\% & 78\% & 71\% & -- & 87\% & 82\% & 72\% & 72\% & 69\% \\
    100 (6) & 85\% & 67\% & 77\% & 78\% & 70\% & 82\% & -- & 79\% & 75\% & 72\% & 66\% \\
    100 (7) & 85\% & 66\% & 72\% & 74\% & 70\% & 80\% & 86\% & -- & 75\% & 74\% & 69\% \\
    100 (8) & 85\% & 62\% & 67\% & 72\% & 66\% & 75\% & 79\% & 81\% & -- & 65\% & 62\% \\
    100 (9) & 85\% & 67\% & 78\% & 76\% & 67\% & 76\% & 86\% & 86\% & 77\% & -- & 74\%\\
    100 (10) & 77\% & 64\% & 73\% & 74\% & 63\% & 77\% & 78\% & 71\% & 64\% & 66\% & -- \\
    \bottomrule
  \end{tabular}
\end{table*}
We evaluated training the model on CIFAR-10 and testing it on CIFAR-100 or vice versa. To provide a similar dataset to CIFAR-10, we select data corresponding to 10 classes of CIFAR-100. We provide evaluations for a comprehensive selection of CIFAR-100 classes in Table \ref{tab:PI_CIFAR}. Our results show an average success rate of 74\% (between 60\% and 87\%) of $PIAttack$ in detecting the presence of a watermark in a given model.

\subsection{Discussion}\label{Discussion}
We provide further discussion of our attacks here and compare our work with other attacks on backdoor-based watermarking schemes. We as well investigate deeper and present evidences that a model has to choose between  a resilient backdoor or high classification accuracy and can not keep both.
\subsubsection{Not the Highest Accuracy}\label{Overlap}
Since we are simulating both the challenger and the attacker in our experiments and desire not allowing overlap in their training dataset, our models effectively have access to half of the training dataset. The limitation prevents our models in MNIST and CIFAR-10 experiments from reaching their highest possible accuracy \cite{CifarRef}. However, despite this limitation, our attacks still successfully remove the embedded watermarks. For CIFAR-100, the non-overlapping requirement results in $10\%$ accuracy loss. Therefore, we decided to allow overlap in training set of the watermarked and the attacker model for this case. 
\subsubsection{Model Selection}
In our experiments, the attacker uses the same model architecture for $\Tilde{M}$ as the owner does for the watermarked target model $\hat{M}$. There are two incentives for this selection: i) as stated by Juuti et al. \cite{PRADA}, higher complexity models increase the prediction accuracy until they are as complex as the source model. Hence, on one hand, the attacker cannot use a model with less capacity than the target model or he will lose accuracy. On the other hand, training a model with higher capacity requires more recourse, e.g more number of queries. This discourages the attacker from training a model with higher capacity than the target mode. Therefore, the target model is the optimal point for the attacker. ii) Watermark removal in our attacks depends on learning the main classification, and not learning the intended misclassification for a trigger set which has a different distribution than the main data. As model resemblance likely increases the chance of \emph{backdoors transfer}, we use the same model for the attacker as the owner to increase the chance of watermark retention to the highest possible value, as the worst case for the attacker. 
\subsubsection{Comparison with Watermark Removal Attacks} \label{Comparison}
Hitaji et al.~\cite{Evasion} introduce Ensemble attack and Evasion attack to remove backdoor-based watermarks. The Ensemble attack, steals $n$ models and collects responses from all of them for each query. It then selects the answer that receives the highest vote among the responds from the stolen networks, and provides that as API prediction. In comparison to this attack, our attacks solely require one marked model and produces a \emph{clean} model that can be redistributed. In the Evasion attack in \cite{Evasion}, a detector mechanism blocks the verification of watermarks. The service will return a random class prediction when it suspects a query is a watermark-trigger. This approach will not be effective in backdoor removal if the trigger samples have a similar distribution to the original samples \cite{Blind-WM}. Chen et al. \cite{Chen} remove the backdoor-based watermark by fine-tuning the model, arguing that previous results  \cite{usenix} on the resilience of backdoor-based watermarking schemes against fine-tuning was due to the low value of learning rate. Unlike our attacks, that use no labeled data, Chen et al. use a combination of labeled ($20\%-80\%$) and unlabeled data in their attack to remove the watermark. 

\subsubsection{Watermark Retention and Test Accuracy} 
\begin{figure}[h!]
\begin{center}
\subfloat[MNIST]{\includegraphics[width=.25\textwidth]{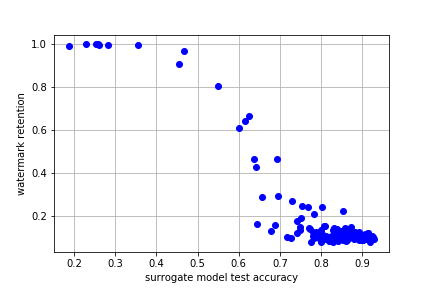}} 
\subfloat[CIFAR-10]{\includegraphics[width=.25\textwidth]{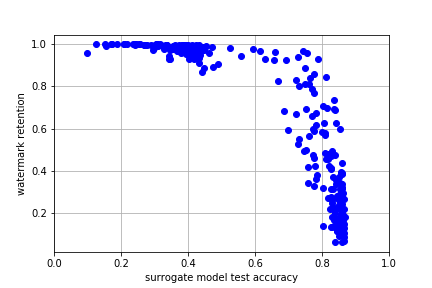}}\hfill
\caption{Watermark retention VS test accuracy of the watermarked model $\hat{M}$, as the number of epochs increases}
\label{fig:Ret_Acc}
\end{center}
\end{figure}

In addition to the successful watermark removal by our attacks, we observed another important result in our experiments. In our black-box attack, we applied the model stealing to a fully trained model $\hat{M}$ with embedded watermarks. To investigate the success of model stealing over partially trained models, we repeatedly applied the attack to the marked model $\hat{M}$ during its training. The resulting test accuracy and watermark retention of the stolen model $\Tilde{M}$ are plotted  in Figure~\ref{fig:Ret_Acc} for both MNIST and CIFAR-10 data sets. 
The attack reaches lower watermark retention if $\hat{M}$ achieves higher test accuracy. This increase in the model accuracy is a result of overcoming the underfitting by increasing the number of epochs during training $\hat{M}$. Consider the Content Embedded watermarking scheme in MNIST dataset. We know that the all elements in the watermark set, which has the size of $10\%$ of the training data, are mapped to a pre-defined fixed class, e.g. class is $2$. Assuming a balanced labeling for the rest of the data,  
the class $2$ is mapped to $20\%$ of the training data in total, resulting in an overall unbalanced class coverage. An underfitted model $\Hat{M}$ transfers this bias to the stolen network $\Tilde{M}$ as well and make it more resilient to watermark removal. 
Clearly, our black-box attack is successful against any $100\%$ accurate model. Hence, watermark retention can only be introduced by inaccuracies in the model.  Our results indicate, however, that these inaccuracies need to be quite significant to result in successful watermark retention.

\section{Related Work}\label{Related_Work}
\subsection{White-box Watermarking Schemes}
The first watermarking scheme for deep neural networks was introduced by Uchida et al.~\cite{uchida}. They propose a white-box watermarking by embedding watermark into the parameters of the DNN model during the training process. Recently, by analyzing the statistical distribution of these models, Wang and Kerschbaum \cite{Florian} presented an attack that detects the watermark and removes it by overwriting. In another watermarking attempt in white-box setting, DeepMarks~\cite{Deepmarks} embeds  a   binary  code-vector  in the  probabilistic  distribution  of  the  weights while preserving accuracy. However, in addition to being susceptible to statistical attacks, white-box watermarking methods \cite{uchida,uchida2,Deepmarks,Tianhao} suffer from application constraints, since access to all the model parameters is not always guaranteed. 
\subsection{Black-box Watermarking Schemes}
The demand in protecting neural networks that are solely accessible through a remote API, has made a tangible shift in DNN watermarking research as well~\cite{comparison}. DeepSigns~\cite{deepsigns} embeds watermarks in the probability density function of the activation set of the target layer and  introduces two versions of the framework to provide watermarking in both white-box and black-box setting. 
In two other approaches \cite{Adversarial, chen2019blackmarks}, the authors use adversarial examples in a zero-bit watermarking algorithm to enable extraction of the watermark without requiring model parameters. This approach however, requires limitation on transferability of the utilized adversarial examples across other networks. Backdoor-based watermarking \cite{usenix,asiaccs,guo,namba2019robust,deepstego} -- as investigated in this paper -- is another recent line of work that aims at watermarking DNN models in the black-box setting. In this approach, a secret trigger set and its pre-defined labels are fed to the model during training process to protect model ownership. 
\subsection{Backdoor Removal}
Since backdooring a neural network may also impose other threats, identifying and removing them has gained attention in research. However, typically such systems as presented in \cite{liu2017neural, chou2018sentinet, gao2019strip} are intended to be employed alongside the neural network. They are tasked only to fend off attempts of actively using the embedded backdoor, which is not applicable for the scenario of attacking watermarking schemes since the trigger set is never released. There are few schemes \cite{Neural-Cleanse,Prune} that do not require access to the trigger set at any time. The Neural-Cleanse approach \cite{Neural-Cleanse} first detects whether a backdoor exists in the model by checking how many pixels in the input image should be modified for the prediction to change to another class. When there is one such consistent small modification for many benign inputs, it is assumed to be a backdoor and then it undergoes reverse-engineering and mitigation processes. This approach only works with backdoors that are restricted to a relatively small patch of the image, which is not the case for all the watermark types we consider in this work. The Fine-Pruning approach \cite{Prune} removes backdoors by pruning redundant
neurons that are less useful for the main classification; which results in drops in the model accuracy for some models \cite{Knockoff}. 
\subsection{Model Extraction}
In addition to a white-box attack for removing backdoor-based watermarks, we propose a black-box attack inspired by model extraction attacks in DNNs. These attacks has been used in several proposals for stealing properties such as hyperparameters \cite{Steal-Hyper}, architecture \cite{Reverse} and membership inference \cite{Membership} from machine learning models in black-box setting.
Tramer et al. used model stealing to present a generic equation-solving attack for models with a logistic output layer and a path-finding algorithm for decision trees. They steal model parameters \cite{Steal} of a neural network using a budget proportional to the model parameters, while assuming knowledge of the model architecture, hyperparameters, and training strategy. Papernot et al. extract a model specifically for forging transferable non-targeted adversarial examples \cite{Steal2} for shallow networks. Inspired by the previous two works, PRADA\cite{PRADA} proposes a general approach for stealing a model efficiently to serve as a surrogate to craft adversarial examples; while requiring partial information about the model. Orekondy et al. \cite{Knockoff} propose functionality stealing with less assumptions, they query the target model by random images from a different distribution than the model's and actively learn on the posterior probability vectors for queries. We use a similar approach to \cite{Knockoff} in our black-box attack. 
\subsection{Removing Backdoor-based watermarks}
In a different approach to remove backdoor-based watermarks, Hitaji et al. introduce ensemble and evasion attacks \cite{Evasion}. In Section \ref{Comparison}, we discussed the limitations of their attacks and how their proposals are essentially different from ours. Chen et al. \cite{Chen} apply fine-tuning to the watermarked model using a combination of labeled and unlabeled data to remove the watermark. Our attacks break the security of backdoor-based watermarking schemes without the need to access labeled data, multiple models, or evading the queries that aim watermark verification. 
\section{Conclusion}
We present three attacks on the recent backdoor-based watermarking schemes in deep neural networks: i) black-box attack, ii) white-box attack, and iii) property inference attack. The watermark in the targeted model could take any of the forms: i) a logo, or an embedded content, ii) a pre-specified noise pattern, or iii) abstract images. Our black-box and white-box attacks do not require access to the trigger set, the ground truth function of the watermarked model or any labeled data. This saves enormous amount of time and resources in preparing the training data. Our attacks use limited number of inputs $(20000-50000)$ from the publicly known domain of the marked model, and query the model for labels. They remove the models' watermark successfully with negligible drop on the classification accuracy. Our black-box approach achieves these goals with minimum access requirements and by solely exploiting the models' classification labels. However, granting more information (e.g.~the marked model's parameters) enables us to devise a white-box attack that is more efficient and accurate than our black-box attack. In addition to our two watermark removal attacks, we proposed a property inference attack that can distinguish a watermarked model from an unmarked one. This attack provides us with a powerful tool, to first recognize watermarked models and then apply our watermark removal attacks on. Second, it benefits our attacks by confirming when the watermark is fully removed, this can help in setting an accurate stop time to further improve the attacks' efficiency. We do not suggest to trust a maliciously backdoored model where the backdoor has been removed using our white-box attack on watermarks; for any -- even minuscule -- retention of the backdoor could be critical.

\bibliographystyle{plain}
\bibliography{references}

\end{document}